\documentclass{article}

\usepackage[preprint]{neurips_2026}
\usepackage[utf8]{inputenc}
\usepackage[T1]{fontenc}
\usepackage{hyperref}
\usepackage{url}
\usepackage{booktabs}
\usepackage{amsfonts}
\usepackage{amsmath}
\usepackage{amssymb}
\usepackage{nicefrac}
\usepackage{xcolor}
\usepackage{graphicx}
\usepackage{multirow}
\usepackage{enumitem}
\usepackage{booktabs}
\usepackage{colortbl}
\usepackage{xcolor}
\usepackage{array}
\usepackage{makecell}
\usepackage{longtable}
\hypersetup{hidelinks}
\usepackage{wrapfig}
\usepackage{placeins}
\usepackage{tabularx}

\usepackage{xurl} % 让 \path 自动断行更好

\newcolumntype{C}{>{\centering\arraybackslash}X}
\title{PULSE: Identifying Demonstration-Utility Features with Sparse Autoencoders}

\author{
  {\small\bfseries
    Chenduo Hao\textsuperscript{1,\textdagger},
    Chuanbao Gao\textsuperscript{1,\textdagger},
    Pinjun Zeng\textsuperscript{1},
    Jingze Zhu\textsuperscript{1},
    Chonghan Liu\textsuperscript{2},
    Zidong Liu\textsuperscript{3},
    Xu Yang\textsuperscript{1,*}
  }\\[0.2em]
  \textsuperscript{1}Southeast University\\
  {\small\ttfamily\{220242441,220242345,213242848,220242456,101013120\}@seu.edu.cn}\\
  \textsuperscript{2}University of California, Los Angeles\\
  {\ttfamily khazzz1c@gmail.com}\\
  \textsuperscript{3}School of Architecture, The University of Texas at Austin\\
  {\ttfamily zidong.liu.22@utexas.edu}
}
\usepackage{tabularx}
\usepackage{ragged2e}

\begin{document}

\maketitle

\begingroup
\renewcommand{\thefootnote}{\textdagger}
\footnotetext{Chenduo Hao and Chuanbao Gao contributed equally.}
\endgroup
\begingroup
\renewcommand{\thefootnote}{*}
\footnotetext{Xu Yang is the corresponding author.}
\endgroup

\begin{abstract}
In-context learning is highly sensitive to demonstration choice, yet most
methods select demonstrations using external query--demonstration
similarity. Such criteria can miss model-specific signals: Similar demonstrations may activate different internal features and downstream behaviors. We introduce \textbf{PULSE}
(\textbf{P}aired \textbf{U}tility \textbf{L}ocalization over
\textbf{S}parse \textbf{E}ncodings), an SAE-based framework for identifying
model-internal features associated with demonstration utility and using them
for demonstration selection. Using a small labeled discovery set, PULSE samples
candidate demonstration sets, measures their zero-shot-relative utility under
the target model, and scores SAE features by how their activation differences
align with utility differences. The top positive and negative coordinates form
a sparse utility-localization vector. We use this vector in two complementary
ways: as a signed score for controlled complete-set ranking, and as
\textbf{PULSE-Retriever}, which converts its magnitude into a feature-relevance
mask for scalable pool-scale retrieval. Across classification, generation, and
reasoning benchmarks, PULSE-Retriever improves over the strongest baseline by
2--3 accuracy points, 0.6--0.9 BLEU-4, and 3.2 exact-match points,
respectively, while controlled ranking validates the identified features
encode a predictive set-level utility signal. Feature inspection and cross-dataset experiments suggest that the identified
features capture task-relevant, dataset-conditioned patterns, yet retain
utility signals that partially transfer across datasets. Our code is available at the
\href{https://github.com/aohenuo/PULSE}{GitHub repository}.
\end{abstract}

\section{Introduction}

In-context learning (ICL) enables large language models (LLMs) to adapt to new tasks at inference time by conditioning on a small set of demonstrations, without any parameter updates \citep{brown2020language}. Despite its practical success, ICL is notoriously sensitive to how demonstrations are configured: for the same query, some demonstrations substantially improve performance while others offer little benefit or even degrade it \citep{zhao2021calibrate, min2022rethinking, yang2023not}. This sensitivity makes demonstration selection a central problem.

Early work observed that demonstrations similar to the query often work better, motivating retrieval methods based on lexical or semantic proximity \citep{liu2022makes, luo2023dr}. While useful, these methods capture only superficial similarity and leave much of the variation in demonstration utility unexplained. Later approaches introduced learnable retrievers trained from richer supervision, including language-model feedback or ranking signals \citep{rubin2022learning, li2023unified}, and more recent methods optimize entire demonstration sets for diversity, coverage, or set-level quality \citep{levy2023diverse, ye2023compositional, yang2023representative, qin2024context, wang2025demonstration}. The success of these methods suggests that good demonstrations depend on latent, non-obvious signals beyond superficial similarity.

\begin{figure}[t]
    \centering
    \includegraphics[width=\linewidth]{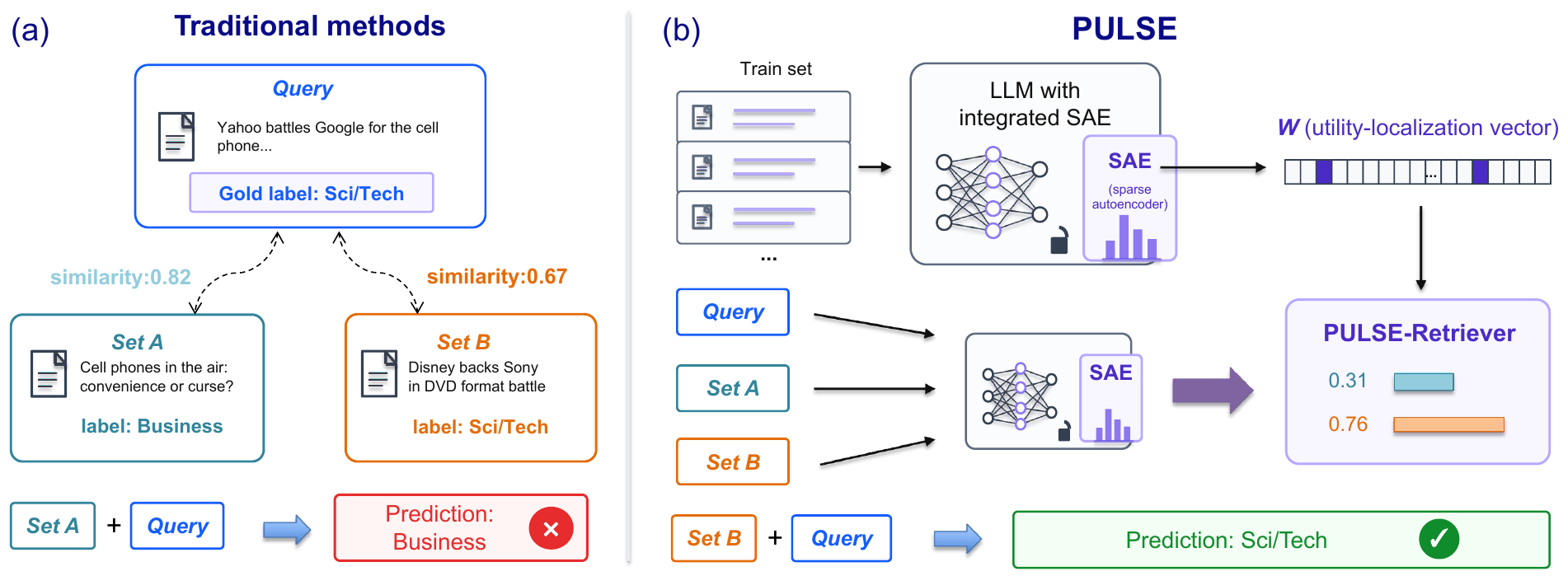}
    \caption{
    Overview of the motivation and workflow of PULSE.
    Traditional methods still study these signals mainly at the level of query--demonstration matching. In contrast, PULSE identifies utility-related internal features and uses them for demonstration selection.
    }
    \label{fig:pulse_intro}
    \vspace{-2.2\baselineskip}
\end{figure}

However, as illustrated in Figure~\ref{fig:pulse_intro}(a), existing methods study these signals primarily at the level of query–demonstration matching, but do not ask which internal features of the model give rise to those signals. This distinction matters because recent work shows that different LLMs exhibit different preferences over demonstrations \citep{peng2024revisiting,wang2024mdr}, utility cannot be determined solely by the external query--demonstration relation, it also depends on which internal features those demonstrations activate in the model. Mechanistic studies support this interpretation: intervening on a small fraction of in-context heads can substantially change ICL performance \citep{akyurek2022learning,olsson2022context,yu2024large}. Taken together, these findings suggest that the key question is not only which demonstrations match the query, but also which internal features of the model make demonstrations useful.

The challenge, then, is that such utility-related features are difficult to isolate in dense activations. Neurons and activation directions are often polysemantic, so signals related to demonstration utility are entangled with many other factors in raw representations \citep{cunningham2023sparse}. Sparse autoencoders (SAEs) provide a natural tool for addressing this challenge: prior work shows that SAE-style decompositions can recover sparse, interpretable features and can now be scaled and deployed broadly across models and layers \citep{cunningham2023sparse, bricken2023monosemanticity, gao2024scaling, lieberum2024gemma,he2024llama}. This creates an opportunity to move from black-box demonstration scoring to explicit identification of the internal features associated with useful demonstrations.

Inspired by this, we propose \textbf{PULSE} (\textbf{P}aired \textbf{U}tility
\textbf{L}ocalization over \textbf{S}parse \textbf{E}ncodings), an SAE-based
framework that identifies utility-related internal features and uses them for demonstration selection, as
shown in Figure~\ref{fig:pulse_intro}(b). For each query, PULSE samples
multiple candidate demonstration sets and forms pairs with
different downstream utilities. It then identifies SAE features whose
activation differences consistently track these utility differences, yielding
a sparse utility-localization vector over SAE features. This vector emphasizes
internal features predictive of demonstration utility while suppressing
variation unrelated to ICL utility.

We then design a PULSE-Ranking experiment to test whether
the identified features encode genuine internal utility signals. For each
evaluation query, we rank randomly sampled demonstration sets
by their alignment with the sparse utility-localization vector. Top-ranked
sets consistently outperform lexical, embedding-based, and SAE-space
baselines, confirming that the identified features distinguish useful
contexts. Building on this, we design \textbf{PULSE-Retriever}, a pool-scale
demonstration retriever to select demonstrations for each query. Because
exhaustively scoring all $k$-shot subsets of a large support pool is
combinatorially infeasible, PULSE-Retriever uses the sparse utility-localization
vector as a feature-relevance mask, combining utility-weighted SAE similarity
with standard SAE similarity to retrieve the final context. Experiments across classification, generation, and reasoning tasks validate that this approach improves demonstration selection across different datasets, shot settings, and backbone models.
Furthermore, we explore what kinds of factors these identified features
capture, and investigate the cross-task transferability of these
features.

Overall, our main contributions are as follows:\begin{itemize}[leftmargin=1.2em, itemsep=0.2em, topsep=0.2em, partopsep=0pt, parsep=0pt]

    \item We reframe ICL demonstration selection as a model-internal sparse
    feature identification problem, and propose PULSE to identify SAE features
    whose activation shifts are associated with useful demonstration sets.

    \item We show, through the PULSE-Ranking experiment, that the
    identified features contain a predictive utility signal for ranking
    demonstration configurations, and introduce PULSE-Retriever, a pool-scale
    demonstration retriever to select demonstrations.
    
    \item We demonstrate across extensive experiments that PULSE improves
    demonstration selection over lexical, embedding-based, and SAE-space
    baselines. Furthermore, we explore the interpretability and transferability
    of the features identified by PULSE.
\end{itemize}

\section{Related Work}

\noindent\textbf{In-context learning and demonstration selection.}
Large language models can perform in-context learning (ICL) by conditioning on a few demonstrations without parameter updates \citep{brown2020language}, but their performance is highly sensitive to the choice, order, and composition of demonstrations \citep{zhao2021calibrate,min2022rethinking,lu2022fantastically,yang2023not}. Existing selection methods estimate demonstration utility from lexical or semantic similarity \citep{liu2022makes,luo2023dr}, learned retrievers \citep{rubin2022learning,zhang2022active,li2023unified,wu2023self}, or set-level criteria such as diversity, coverage, representativeness, and context quality \citep{levy2023diverse,ye2023compositional,yang2023representative,qin2024context,wang2025demonstration,chen2025enhancing}. These approaches are effective, but they mainly operate on external query--demonstration or demonstration-set signals.

Recent work shows that demonstration utility can be model-dependent: effective demonstrations may vary across inference models, motivating model-aware scoring or retrieval criteria \citep{peng2024revisiting,wang2024mdr}. Mechanistic analyses further support this view, showing that ICL can depend on specific internal components, such as in-context heads \citep{yu2024large}, and on the balance between learning from demonstrations and retrieving internal knowledge \citep{nafar2025learning}. These findings motivate our shift from external demonstration scoring to model-internal utility identification with SAE features.

\noindent\textbf{Sparse autoencoders for interpretable model features.}
A key obstacle to analyzing model-internal utility is that neural activations
are dense and entangled. Under the superposition hypothesis, many features may
be represented in overlapping activation directions, making neurons or raw
directions difficult to interpret \citep{elhage2022toy,cunningham2023sparse}.
Sparse autoencoders (SAEs) decompose activations into sparse feature
representations that are often more interpretable
\citep{olshausen1997sparse,bricken2023monosemanticity,cunningham2023sparse}. Recent SAE
scaling efforts and public SAE releases across model families further make SAE
features a practical space for studying model-specific behavior
\citep{gao2024scaling,templeton2024scaling,rajamanoharan2024jumping,lieberum2024gemma}.
PULSE uses this space to identify features whose activation shifts distinguish
higher-utility from lower-utility demonstration sets, and then uses the
resulting utility-localization vector for demonstration selection.

\section{Method}
\label{sec:method}

\begin{figure}[t]
    \centering
    \includegraphics[width=\linewidth]{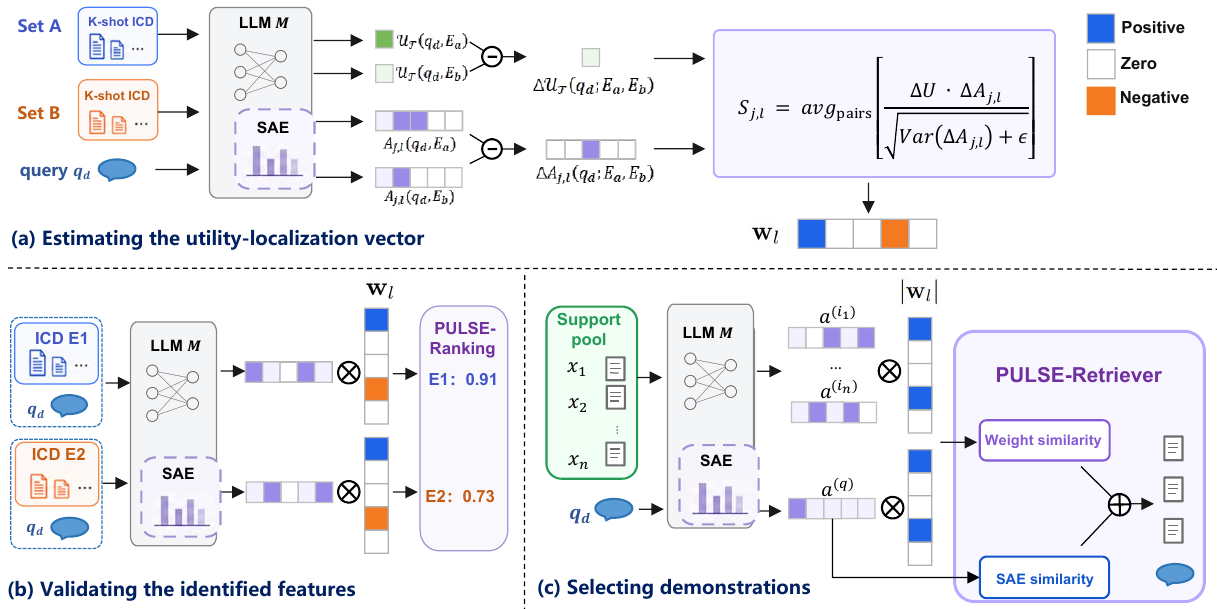}
    \caption{\textbf{Overview of PULSE.}
Here, $k$-shot ICD denotes a $k$-example in-context demonstration set.
(a) PULSE estimates a utility-localization vector by associating utility
differences $\Delta \mathcal{U}$ with SAE activation differences $\Delta A$
across paired demonstration sets.
(b) PULSE validates the identified features by ranking pre-sampled complete
demonstration sets in the PULSE-Ranking experiment.
(c) PULSE selects demonstrations using PULSE-Retriever, which uses the
utility-localization vector as a feature-relevance mask for scalable
pool-scale retrieval.}
    \label{fig:PULSE_method}
    \vspace{-10pt}
\end{figure}

We propose PULSE, a framework that first estimates the utility-localization vector to identify utility-relevant SAE features, then validates the identified features through the PULSE-Ranking experiment, and finally uses PULSE-Retriever to select demonstrations.
PULSE consists of three components, as shown in
Figure~\ref{fig:PULSE_method}: estimating a sparse utility-localization vector
from paired demonstration sets, validating the identified features with
PULSE-Ranking experiment, and selecting demonstrations by PULSE-Retriever.

\noindent\textbf{Setup.}
Let $\mathcal{S}$ be the supervised support pool, where each example is
$(x_i,z_i)$. Here $z_i$ is the task target: a class label for classification
(drawn from label set $\mathcal{Y}$), a reference sentence for CommonGen,
or a supervised solution sequence for GSM8K. For classification, we write
$y^\star(x)\in\mathcal{Y}$ for the gold label of input $x$. For an input $x$,
$q_x$ denotes the prediction query with the target hidden. A $k$-shot
demonstration set is
$E=\{(x_1,z_1),\ldots,(x_k,z_k)\}\subset\mathcal{S}$, and
$P_{\mathcal{T}}(E,q_x)$ denotes the task prompt containing $E$ followed by
$q_x$; $P_{\mathcal{T}}(\varnothing,q_x)$ denotes the corresponding zero-shot
prompt.

Let $M$ be the backbone language model with output distribution
$p_M(\cdot\mid\cdot)$, and a pretrained SAE attached to residual-stream
layer $l$. We pass the prompt through $M$ and the SAE, mean-pool SAE
activations over prompt tokens, and denote the sparse feature
vector by $\mathbf{A}_l(q_x,E)=(A_{j,l}(q_x,E))_j$.

\noindent\textbf{Utility definition.}
PULSE uses labels only in a small discovery set to estimate which SAE features
are associated with demonstration utility. For classification tasks, we use
the gold-label margin:
\begin{equation}
\label{eq:margin}
\mathcal{G}_{\mathrm{cls}}(q_x,E)
=
\log p_M(y^\star(x)\mid P_{\mathcal{T}}(E,q_x))
-
\max_{y\in\mathcal{Y},\,y\neq y^\star(x)}
\log p_M(y\mid P_{\mathcal{T}}(E,q_x)).
\end{equation}

For generative tasks, we use the length-normalized teacher-forced likelihood
of the target sequence $z^\star=(z^\star_1,\ldots,z^\star_T)$:
\begin{equation}
\label{eq:gen_score}
\mathcal{G}_{\mathrm{gen}}(q_x,E)
=
\frac{1}{T}
\sum_{t=1}^{T}
\log p_M
\left(
z^\star_t
\mid
P_{\mathcal{T}}(E,q_x), z^\star_{<t}
\right).
\end{equation}
This utility proxy is used only to estimate the utility-localization vector; final performance is
evaluated with the standard task metric. We define zero-shot-relative utility as
\begin{equation}
\label{eq:relative_utility}
\mathcal{U}_{\mathcal{T}}(q_x,E)
=
\mathcal{G}_{\mathcal{T}}(q_x,E)
-
\mathcal{G}_{\mathcal{T}}(q_x,\varnothing),
\end{equation}
where $\mathcal{G}_{\mathcal{T}}$ is instantiated as
$\mathcal{G}_{\mathrm{cls}}$ for classification and
$\mathcal{G}_{\mathrm{gen}}$ for generation.

\noindent\textbf{Estimating the utility-localization vector.}
We estimate $\mathbf{w}_l$ using a labeled discovery set
$\mathcal{D}_{\mathrm{disc}}$ sampled from the training split and disjoint from
evaluation queries.
Each discovery example $d=(x_d,z_d)$ is treated as a query $q_d=q_{x_d}$.
For each $q_d$, we sample $N$ candidate demonstration sets from
$\mathcal{S}$, excluding $d$ itself if present, and form within-query pairs
$(E_a,E_b)$. Within-query pairing controls for query-specific difficulty, so
$\Delta \mathcal{U}_{\mathcal{T}}$ mainly reflects relative demonstration-set
utility rather than differences across queries. For each pair, we compute the utility difference
\begin{equation}
\label{eq:delta_utility}
\Delta\mathcal{U}_{\mathcal{T}}(q_d;E_a,E_b)
=
\mathcal{U}_{\mathcal{T}}(q_d,E_a)
-
\mathcal{U}_{\mathcal{T}}(q_d,E_b),
\end{equation}
and the SAE activation difference
\begin{equation}
\label{eq:delta_activation}
\Delta A_{j,l}(q_d;E_a,E_b)
=
A_{j,l}(q_d,E_a)
-
A_{j,l}(q_d,E_b).
\end{equation}

PULSE scores each SAE feature by its activation-normalized association with utility differences.
Let $n_{\mathrm{pairs}} = |\mathcal{D}_{\mathrm{disc}}| \cdot \binom{N}{2}$ denote the total number of unordered pairs, and let $\mathrm{Var}(\Delta A_{j,l})$ denote the variance of $\Delta A_{j,l}$ computed globally over all pairs. The feature score is
\begin{equation}
\label{eq:pulse_feature_score}
S_{j,l}
=
\frac{
\sum_{(q_d,E_a,E_b)}
\Delta \mathcal{U}_{\mathcal{T}}(q_d;E_a,E_b)\,
\Delta A_{j,l}(q_d;E_a,E_b)
}{
n_{\mathrm{pairs}}
\sqrt{\mathrm{Var}(\Delta A_{j,l})+\varepsilon}
},
\end{equation}
where $\varepsilon$ is a small stability constant. This score measures the
normalized alignment between $\Delta A_{j,l}$ and $\Delta \mathcal{U}_{\mathcal{T}}$.
We define $\mathbf{w}_l$ coordinate-wise by setting
$w_{j,l}=S_{j,l}$ for features whose scores are among the $K^+$ largest
positive or $K^-$ most negative values, and $w_{j,l}=0$ otherwise. \\
\noindent\textbf{Validating the identified features.}
To test whether the identified features encode an internal utility signal, we design a PULSE-Ranking experiment. For each evaluation query, all methods rank the same
pre-sampled complete demonstration sets. PULSE ranks a candidate set $E$ by
the alignment between its complete-context activation shift and the signed
utility-localization vector:
\begin{equation}
\label{eq:set_score}
s_{\mathrm{rank}}(q_x,E)
=
\mathbf{w}_l^{\top}
\bigl[
\mathbf{A}_l(q_x,E)-\mathbf{A}_l(q_x,\varnothing)
\bigr].
\end{equation}
A higher score indicates that the context activates features associated with
higher utility. This experiment is a controlled validation of
the identified features, not the practical pool-scale retriever.

\noindent\textbf{Selecting demonstrations.}
Equation~\eqref{eq:set_score} scores complete demonstration sets with the
signed utility-localization vector, but applying it to all
$\binom{|\mathcal{S}|}{k}$ candidate sets is combinatorially infeasible. We
therefore design PULSE-Retriever as a scalable proxy that ranks
individual support examples before constructing the final $k$-shot context.

To adapt the set-level signal to pointwise retrieval, we use
$|\mathbf{w}_l|$ as a feature-relevance mask. The sign of $\mathbf{w}_l$ is
meaningful for complete-set scoring, where it records the direction of
association between full-context activation shifts and utility differences.
However, it is not calibrated as an individual-candidate preference direction.
Thus, $|\mathbf{w}_l|$ identifies which SAE coordinates are utility-informative
while avoiding an unsupported pointwise interpretation of the sign. Let
$\mathbf{a}^{(q)}=\mathbf{A}_l(q_x,\varnothing)$ and
$\mathbf{a}^{(i)}=\mathbf{A}_l(q_{x_i},\varnothing)$ be zero-shot SAE
encodings of the query and candidate input, both with targets hidden. The
PULSE-masked similarity is
\begin{equation}
\label{eq:weighted_cosine}
s_{\mathrm{mask}}(q_x,x_i)
=
\cos\!\bigl(
|\mathbf{w}_l|\odot\mathbf{a}^{(q)},\;
|\mathbf{w}_l|\odot\mathbf{a}^{(i)}
\bigr).
\end{equation}

We blend this masked similarity with standard SAE-space cosine similarity:
\begin{equation}
\label{eq:blend}
s_{\mathrm{blend}}(q_x,x_i)
=
(1-\beta)\,\hat{s}_{\mathrm{mask}}(q_x,x_i)
+
\beta\,\hat{s}_{\mathrm{cos}}(q_x,x_i),
\end{equation}
where
$s_{\mathrm{cos}}(q_x,x_i)=\cos(\mathbf{a}^{(q)},\mathbf{a}^{(i)})$,
$\hat{\cdot}$ denotes per-query $z$-score normalization, and
$\beta\in[0,1]$. The final $k$-shot context is greedily selected from a
top-$L$ shortlist with a redundancy penalty in SAE space. Thus,
PULSE-Retriever uses the identified internal features as a scalable retrieval
proxy, rather than as an exact decomposition of complete-context utility.

\section{Experiments}
\label{sec:setup}
\subsection{Experiment Setup}
\noindent\textbf{Models and SAEs.}
Because PULSE identifies utility-related features in SAE feature space, we
evaluate it on models with publicly available SAEs: Gemma2-2B \cite{team2024gemma} and
Llama3.1-8B \citep{grattafiori2024llama}. Unless otherwise specified, experiments use both backbones;
GSM8K is evaluated only with Llama3.1-8B.

\noindent\textbf{Datasets and metrics.}
We evaluate on six benchmarks spanning classification, generation, and
reasoning. For classification, we use AGNews \cite{zhang2015character} (4-class topic classification),
REST14 and LAP14 (3-class aspect-based sentiment) \cite{pontiki-etal-2014-semeval}, and EMOC (4-class emotion
recognition) \cite{chatterjee2019semeval}, reporting classification accuracy. For generation, we use
CommonGen \cite{lin2020commongen}, reporting BLEU-4. For reasoning, we use
GSM8K \cite{cobbe2021training}, reporting exact match accuracy. Across all
tasks, utility-related features are identified only from the training split,
and held-out test data are used for ICL evaluation. 

\noindent\textbf{Protocol.}
PULSE estimates $\mathbf{w}_l$ from
$|\mathcal{D}_{\mathrm{disc}}|=64$ discovery queries sampled from the training
split and disjoint from evaluation queries, with $N=32$ candidate
demonstration sets per query. We evaluate the identified features in two
settings: \textbf{PULSE-Ranking experiment}, a controlled validation where all
methods rank the same pre-sampled complete $k$-shot sets, and
\textbf{PULSE-Retriever}, a practical pool-scale retriever that uses the same
features as a feature-relevance mask. We further evaluate PULSE-Retriever on
CommonGen and GSM8K for generation and reasoning. Unless otherwise specified,
$K^{+}=K^{-}=512$ for Gemma2-2B and $K^{+}=K^{-}=2048$ for Llama3.1-8B;
PULSE-Retriever uses $L=50$ and $\lambda_r=0.3$. All hyperparameters are calibrated on a held-out validation split and then 
fixed before evaluation on the test set.

\noindent\textbf{Baselines.}
We compare PULSE against five baselines covering random, lexical, dense-embedding, SAE-space, and diversity-aware selection: \textbf{Random} (uniform sampling); \textbf{Lex-Sim} (Jaccard lexical similarity); \textbf{SBERT} \cite{reimers2019sentence}, cosine similarity over \texttt{all-MiniLM-L6-v2} \cite{wang2020minilm} embeddings; \textbf{KNN-SAE}, cosine similarity in SAE feature space; and \textbf{CEIL}~\citep{ye2023compositional}, DPP-based diversity selection over pretrained embeddings. Since PULSE uses a small labeled discovery set, Appendix~\ref{app:epr_budget} further compares against an \textbf{EPR}~\citep{rubin2022learning} learned retriever with a larger feedback budget, isolating whether the gains come from SAE-based utility identification rather than labeled feedback alone.

\subsection{Main Results}
\label{sec:results}

\noindent\textbf{PULSE-Ranking Experiment.}
\label{sec:detector}
To evaluate whether the utility-localization vector learned by PULSE captures
a set-level utility signal, we design a PULSE-Ranking experiment, with results shown in Table~\ref{tab:detector}. Across $32$
dataset--shot--backbone cells, PULSE-Ranking achieves the best or
tied-best accuracy in $30$ cases, outperforming lexical similarity, sentence
embeddings, SAE nearest neighbors, and diversity-aware selection. Because all
methods rank identical pre-sampled candidate configurations, this directly
shows that PULSE provides a stronger score for identifying high-utility
demonstration sets. The gain over KNN-SAE is substantial---$3.89$ points on
Gemma2-2B and $3.58$ points on Llama3.1-8B---indicating that the improvement
stems from the utility-related SAE feature weights identified by PULSE, not
merely from using SAE representations. These results validate that the identified features encode a predictive
set-level utility signal: they distinguish higher-utility from lower-utility
complete demonstration configurations under a controlled ranking protocol.

% To test whether PULSE discovers a useful set-level utility signal, we evaluate it on the candidate-set scoring task. Table~\ref{tab:detector} reports the results on Gemma2-2B and Llama3.1-8B. PULSE achieves the best or tied-best accuracy in $30$ of the $32$ dataset--shot--backbone cells. Since all methods rank the same pre-sampled candidate configurations, this result directly shows that PULSE provides a stronger signal for identifying high-utility demonstration sets than lexical similarity, sentence embeddings, SAE nearest neighbors, and diversity-aware selection.

% Compared with KNN-SAE, PULSE improves average accuracy by $3.89$ points on Gemma2-2B and $3.58$ points on Llama3.1-8B. This indicates that the gain does not come merely from using SAE representations, but from learning utility-relevant SAE feature weights. Overall, these results validate the first role of PULSE: the learned utility-localization vector can distinguish between stronger and weaker complete demonstration configurations.

\begin{table}[t]
\centering
\scriptsize
\setlength{\tabcolsep}{1.8pt}
\caption{\textbf{Candidate-set scoring accuracy (\%) on Gemma2-2B and Llama3.1-8B.}
Downstream accuracy after using each score to select the top-1 candidate
demonstration set. The Avg. columns report the average across four datasets
under each shot setting. }
\label{tab:detector}
\resizebox{\textwidth}{!}{
\begin{tabular}{ll*{20}{c}}
\toprule
Backbone & Method
& \multicolumn{4}{c}{AGNews}
& \multicolumn{4}{c}{REST14}
& \multicolumn{4}{c}{LAP14}
& \multicolumn{4}{c}{EMOC}
& \multicolumn{4}{c}{Avg.} \\
\cmidrule(lr){3-6}
\cmidrule(lr){7-10}
\cmidrule(lr){11-14}
\cmidrule(lr){15-18}
\cmidrule(lr){19-22}
& & 1 & 2 & 4 & 8
& 1 & 2 & 4 & 8
& 1 & 2 & 4 & 8
& 1 & 2 & 4 & 8
& 1 & 2 & 4 & 8 \\
\midrule
\multirow{6}{*}{Gemma2-2B}
& Random
& 40.62 & 66.99 & 75.00 & 77.93
& 60.94 & 65.82 & 73.83 & 73.05
& 54.21 & 65.44 & 76.03 & 73.87
& 44.92 & 53.52 & 54.69 & 60.16
& 50.17 & 62.94 & 69.89 & 71.25 \\
& Lex-Sim
& 43.75 & 68.75 & 77.34 & 82.03
& 66.41 & 72.46 & 78.71 & 75.39
& 64.79 & 70.84 & 78.40 & 77.75
& 52.93 & 54.88 & 56.64 & 60.94
& 56.97 & 66.73 & 72.77 & 74.03 \\
& SBERT
& 55.47 & 77.15 & 78.12 & 82.23
& 60.74 & 69.14 & 76.17 & 73.83
& 66.52 & 71.49 & 75.81 & 77.11
& 51.56 & 55.86 & 56.05 & 61.33
& 58.57 & 68.41 & 71.54 & 73.62 \\
& CEIL
& 55.47 & 71.48 & 78.12 & 78.12
& 60.74 & 69.73 & 76.76 & 75.59
& 66.52 & 70.41 & 79.70 & 78.19
& 51.56 & \textbf{56.64} & 56.25 & 60.94
& 58.57 & 67.06 & 72.71 & 73.21 \\
& KNN-SAE
& 54.69 & 77.54 & 76.95 & 81.84
& 63.67 & 69.53 & 78.12 & 76.17
& 65.87 & 72.35 & 79.05 & 76.24
& 48.63 & 52.93 & 58.01 & 60.55
& 58.22 & 68.09 & 73.03 & 73.70 \\
& \textbf{PULSE-Ranking}
& \textbf{57.03} & \textbf{78.12} & \textbf{82.03} & \textbf{82.62}
& \textbf{75.59} & \textbf{77.34} & \textbf{80.47} & \textbf{78.71}
& \textbf{73.43} & \textbf{74.95} & \textbf{81.86} & \textbf{79.48}
& \textbf{54.69} & 55.66 & \textbf{60.94} & \textbf{61.52}
& \textbf{65.18} & \textbf{71.52} & \textbf{76.32} & \textbf{75.58} \\
\midrule
\multirow{6}{*}{Llama3.1-8B}
& Random
& 43.95 & 60.16 & 73.24 & 82.23
& 75.39 & 74.41 & 76.76 & 77.15
& 77.11 & 76.89 & 79.91 & 80.35
& 45.12 & 57.03 & 60.16 & 63.28
& 60.39 & 67.12 & 72.52 & 75.75 \\
& Lex-Sim
& 53.32 & 66.99 & 74.61 & 75.20
& 76.17 & 77.34 & 79.49 & 77.93
& \textbf{78.19} & 78.19 & 79.05 & 79.27
& 50.20 & 58.59 & 60.94 & 62.50
& 64.47 & 70.28 & 73.52 & 73.72 \\
& SBERT
& 59.77 & 74.02 & 78.32 & 81.25
& 75.78 & 74.80 & 78.52 & 78.12
& 77.97 & 78.83 & 78.40 & 79.05
& 55.47 & 60.35 & 63.28 & 64.45
& 67.25 & 72.00 & 74.63 & 75.72 \\
& CEIL
& 59.77 & 75.59 & 78.91 & 82.42
& 75.78 & 76.37 & \textbf{80.08} & 78.71
& 77.97 & 79.05 & 78.83 & 79.70
& 55.47 & 60.74 & 66.60 & 67.77
& 67.25 & 72.94 & 76.10 & 77.15 \\
& KNN-SAE
& 55.66 & 74.41 & 80.47 & 83.20
& 74.61 & 74.61 & 78.12 & 78.52
& 76.67 & \textbf{79.70} & 79.70 & 78.40
& 50.98 & 59.77 & 60.94 & 64.84
& 64.48 & 72.12 & 74.81 & 76.24 \\
& \textbf{PULSE-Ranking}
& \textbf{62.50} & \textbf{82.03} & \textbf{84.38} & \textbf{85.94}
& \textbf{76.56} & \textbf{80.47} & \textbf{80.08} & \textbf{80.27}
& 77.97 & \textbf{79.70} & \textbf{80.13} & \textbf{80.56}
& \textbf{56.25} & \textbf{64.06} & \textbf{67.58} & \textbf{69.34}
& \textbf{ 68.32} & \textbf{76.56} & \textbf{78.04} & \textbf{79.03} \\
\bottomrule
\end{tabular}
}
%\vspace{-20pt}
\end{table}

%\subsection{Pool-Scale Demonstration Retrieval}
\noindent\textbf{Pool-Scale Demonstration Retrieval.}
\label{sec:retriever}
We evaluate the effectiveness of PULSE-Retriever, the pool-scale
demonstration retriever derived from PULSE, with results reported in
Table~\ref{tab:retriever}. Unlike the candidate-set scoring setup, this
setting does not pre-enumerate complete $k$-shot configurations; instead,
each method retrieves individual examples from a large pool to form the final
context. PULSE-Retriever achieves the best accuracy in $25$ out of $32$
dataset--shot--backbone cells, demonstrating that the utility-related features
identified by PULSE can be converted into a practical retrieval signal.

Compared to KNN-SAE, PULSE-Retriever improves average accuracy by $2.93$ points
on Gemma2-2B and $1.99$ points on Llama3.1-8B. These gains are smaller than
 candidate-set scoring, as expected: PULSE-Retriever uses the learned
utility-localization vector as a scalable per-example feature mask rather
than directly scoring complete $k$-shot sets. Nevertheless, the consistent
improvements confirm that model-internal sparse utility features remain useful
for practical demonstration construction. We further verify the stability of PULSE-Retriever under multiple random seeds
in Appendix~\ref{app:seed_stability}. 

% To test whether the same utility signal can support practical retrieval, we evaluate PULSE on pool-scale demonstration retrieval. Table~\ref{tab:retriever} reports the results. Unlike candidate-set scoring, this setting does not enumerate complete demonstration sets in advance. Instead, each method must retrieve individual examples and construct the final $k$-shot context from the pool. PULSE-Retriever achieves the best accuracy in $25$ out of $32$ dataset–shot–backbone cells, showing that the utility signal learned from set-level comparisons can also guide practical retrieval.

% Compared with KNN-SAE, PULSE-Retriever improves average accuracy by $2.93$ points on Gemma2-2B and $1.99$ points on Llama3.1-8B. These gains are smaller than those in candidate-set scoring, which is expected: retrieval uses the learned utility vector as a scalable per-example proxy rather than directly scoring complete $k$-shot configurations. Nevertheless, the consistent improvements demonstrate that model-internal sparse utility features remain useful for realistic demonstration construction.

\begin{table}[t]
\centering
\scriptsize
\setlength{\tabcolsep}{1.8pt}
\caption{\textbf{Pool-scale retrieval accuracy (\%) on Gemma2-2B and Llama3.1-8B.}
Downstream accuracy after retrieving individual examples from the support set
and composing the final $k$-shot context. The Avg. columns report the average
across four datasets under each shot setting. }
\label{tab:retriever}
\resizebox{\textwidth}{!}{
\begin{tabular}{ll*{20}{c}}
\toprule
Backbone & Method
& \multicolumn{4}{c}{AGNews}
& \multicolumn{4}{c}{REST14}
& \multicolumn{4}{c}{LAP14}
& \multicolumn{4}{c}{EMOC}
& \multicolumn{4}{c}{Avg.} \\
\cmidrule(lr){3-6}
\cmidrule(lr){7-10}
\cmidrule(lr){11-14}
\cmidrule(lr){15-18}
\cmidrule(lr){19-22}
& & 1 & 2 & 4 & 8
& 1 & 2 & 4 & 8
& 1 & 2 & 4 & 8
& 1 & 2 & 4 & 8
& 1 & 2 & 4 & 8 \\
\midrule
\multirow{6}{*}{Gemma2-2B}
& Random
& 41.80 & 66.99 & 72.66 & 78.52
& 57.62 & 64.65 & 70.90 & 72.46
& 56.59 & 60.91 & 68.25 & 71.71
& 41.60 & 54.69 & 56.25 & 60.94
& 49.40 & 61.81 & 67.02 & 70.91 \\
& Lex-Sim
& 68.16 & 78.32 & 79.88 & 86.13
& 64.45 & 66.80 & 73.63 & 76.76
& 70.63 & 75.16 & 74.30 & 77.54
& 62.70 & 62.50 & 65.82 & 71.68
& 66.48 & 70.69 & 73.41 & 78.03 \\
& SBERT
& 76.76 & 78.52 & 79.49 & 84.18
& 65.62 & 66.99 & 71.68 & 74.22
& 68.47 & 73.87 & 75.81 & 78.62
& 63.48 & 64.65 & 66.80 & 72.46
& 68.58 & 71.01 & 73.45 & 77.37 \\
& CEIL
& 76.76 & 78.91 & 81.25 & 87.11
& 65.62 & 67.97 & 75.39 & 76.95
& 68.47 & 76.03 & 76.03 & 78.83
& 63.48 & 65.62 & 66.02 & 71.09
& 68.58 & 72.13 & 74.67 & 78.50 \\
& KNN-SAE
& 79.69 & 78.71 & 81.05 & 87.50
& \textbf{66.99} & 65.43 & 75.59 & 75.59
& 70.63 & 76.46 & 75.16 & 77.75
& 50.20 & 60.74 & 66.80 & \textbf{72.46}
& 66.88 & 70.33 & 74.65 & 78.32 \\
& \textbf{PULSE-Retriever}
& \textbf{81.25} & \textbf{82.62} & \textbf{85.94} & \textbf{88.09}
& 64.65 & \textbf{71.68} & \textbf{77.73} & \textbf{77.73}
& \textbf{72.79} & \textbf{77.11} & \textbf{77.32} & \textbf{80.35}
& \textbf{63.87} & \textbf{65.82} & \textbf{68.36} & 72.27
& \textbf{70.64} & \textbf{74.31} & \textbf{77.34} & \textbf{79.61} \\
\midrule
\multirow{6}{*}{Llama3.1-8B}
& Random
& 50.39 & 69.92 & 81.45 & 83.40
& 75.00 & 75.59 & 76.37 & 77.54
& 77.11 & 77.11 & 77.32 & 77.32
& 48.83 & 55.47 & 62.11 & 65.23
& 62.83 & 69.52 & 74.31 & 75.87 \\
& Lex-Sim
& 69.53 & 81.84 & 87.70 & 87.50
& 75.20 & 78.12 & 79.30 & 79.10
& \textbf{77.75} & 79.05 & 80.99 & 81.43
& 61.91 & 69.92 & 62.70 & 67.19
& 71.10 & 77.23 & 77.67 & 78.81 \\
& SBERT
& 80.08 & 80.86 & 86.72 & 87.70
& 75.59 & 80.08 & 80.47 & 80.27
& 71.06 & 80.35 & 81.43 & 81.43
& \textbf{65.23} & 68.95 & 65.23 & 69.34
& 72.99 & 77.56 & 78.46 & 79.68 \\
& CEIL
& 80.08 & 85.16 & 85.55 & 88.48
& 75.59 & 80.66 & 79.88 & 80.47
& 71.06 & 78.83 & 81.64 & 81.21
& \textbf{65.23} & 71.09 & 68.95 & 74.22
& 72.99 & 78.94 & 79.00 & 81.10 \\
& KNN-SAE
& \textbf{81.25} & 84.77 & 86.13 & 88.09
& 75.39 & 76.56 & 80.66 & 81.25
& 68.03 & \textbf{81.86} & \textbf{82.72} & 82.29
& 61.13 & 68.36 & 66.02 & 73.83
& 71.45 & 77.89 & 78.88 & 81.36 \\
& \textbf{PULSE-Retriever}
& 79.88 & \textbf{86.52} & \textbf{88.09} & \textbf{89.84}
& \textbf{77.34} & \textbf{81.45} & \textbf{81.84} & \textbf{81.45}
& 76.03 & 80.56 & 82.07 & \textbf{82.72}
& 64.26 & \textbf{71.29} & \textbf{71.09} & \textbf{75.78}
& \textbf{74.38} & \textbf{79.96} & \textbf{80.77} & \textbf{82.45} \\
\bottomrule
\end{tabular}
}
\vspace{-10pt}
\end{table}

%\subsection{Generalization to Generation and Reasoning}
\noindent\textbf{Generalization to Generation and Reasoning.}
\label{sec:gen_reasoning}
Beyond classification datasets, we further evaluate PULSE-Retriever on
generation (CommonGen) and reasoning (GSM8K) tasks. Unlike the previous
classification tasks, these benchmarks require open-ended generation or
multi-step mathematical reasoning, offering a stronger test of whether the
utility-related features identified by PULSE can support broader demonstration
selection. As reported in Table~\ref{tab:gen-reasoning}, PULSE-Retriever achieves
the best result in all $12$ setting--shot cells: on CommonGen, it improves
over the strongest baseline by $+0.93$ BLEU-4 on Gemma2-2B and $+0.61$ BLEU-4
on Llama3.1-8B when averaged over shots; on GSM8K, it improves by $+3.16$
exact match. These results show the utility-related internal features
identified by PULSE are not limited to classification but also benefit
generation and reasoning tasks.

% We further test whether the utility signal identified by PULSE is limited to
% classification tasks. Table~\ref{tab:gen-reasoning} reports results on
% CommonGen and GSM8K. Unlike the previous four tasks, CommonGen requires
% open-ended generation and GSM8K requires multi-step mathematical reasoning.
% Thus, these benchmarks provide a stronger test of whether PULSE can guide
% demonstration selection beyond standard classification.

% PULSE achieves the best result in all $12$ setting--shot cells. On CommonGen,
% PULSE improves over the strongest baseline by $+0.93$ BLEU-4 on Gemma2-2B and
% $+0.61$ BLEU-4 on Llama3.1-8B when averaged over shots. On GSM8K, it improves
% over the strongest baseline by $+3.17$ EM. These results suggest that the
% identified utility-related internal features are not specific to
% instruction-free classification, but can also support demonstration selection
% for generation and reasoning tasks.

\begin{table*}[t]
\centering
\scriptsize
\setlength{\tabcolsep}{2.6pt}
\renewcommand{\arraystretch}{0.92}
\caption{
Generation and reasoning results. Scores are BLEU-4 for CommonGen and exact match accuracy for GSM8K. 
The Avg. columns report the average over 1-, 2-, 4-, and 8-shot settings. 
}
\label{tab:gen-reasoning}
\begin{tabular*}{\textwidth}{@{\extracolsep{\fill}}lccccccccccccccc@{}}
\toprule
\multirow{2}{*}{Method}
& \multicolumn{5}{c}{CommonGen / Gemma2-2B / BLEU-4}
& \multicolumn{5}{c}{CommonGen / Llama3.1-8B / BLEU-4}
& \multicolumn{5}{c}{GSM8K / Llama3.1-8B/ EM} \\
\cmidrule(lr){2-6}
\cmidrule(lr){7-11}
\cmidrule(lr){12-16}
& 1 & 2 & 4 & 8 & Avg.
& 1 & 2 & 4 & 8 & Avg.
& 1 & 2 & 4 & 8 & Avg. \\
\midrule

Random
& 8.42 & 8.35 & 8.10 & 8.86 & 8.43
& 9.02 & 9.22 & 9.64 & 9.44 & 9.33
& 42.61 & 47.92 & 49.20 & 50.72 & 47.61 \\

Lex-Sim
& 8.22 & 7.74 & 9.20 & 10.56 & 8.93
& 9.68 & 8.91 & 9.19 & 9.16 & 9.24
& 38.29 & 48.67 & 50.04 & 52.31 & 47.33 \\

SBERT
& 8.41 & 8.36 & 9.08 & 9.82 & 8.92
& 9.41 & 9.84 & 10.21 & 9.91 & 9.84
& 39.73 & 49.20 & 43.82 & 52.92 & 46.42 \\

KNN-SAE
& 9.00 & 8.34 & 8.93 & 7.55 & 8.46
& 8.91 & 9.64 & 9.96 & 9.71 & 9.56
& 39.12 & 46.10 & 46.93 & 53.53 & 46.42 \\

CEIL
& 8.41 & 8.62 & 10.07 & 9.24 & 9.09
& 9.41 & 9.68 & 10.01 & 9.47 & 9.64
& 39.73 & 50.42 & 50.80 & 52.08 & 48.26 \\

\textbf{PULSE-Retriever}
& \textbf{9.05} & \textbf{9.34} & \textbf{11.09} & \textbf{10.59} & \textbf{10.02}
& \textbf{9.91} & \textbf{10.91} & \textbf{10.90} & \textbf{10.09} & \textbf{10.45}
& \textbf{43.37} & \textbf{50.72} & \textbf{54.66} & \textbf{56.94} & \textbf{51.42} \\

\bottomrule
\end{tabular*}

\vspace{-0.8em}
\end{table*}

\subsection{Ablation Studies}
\label{sec:ablations}

We ablate the SAE basis and the identified PULSE features in the 4-shot
retriever setting. Additional hyperparameter sensitivity results for the
number of identified features, the sparse--dense blend weight $\beta$, and the
number of discovery queries are reported in Appendix~\ref{app:hyperparam}.

\noindent\textbf{Do SAE features provide a better internal basis?}
% To test whether SAE features specifically provide a better internal basis for
% PULSE, we replace them with two alternative internal representations while
% keeping the paired utility-discovery objective and the rest of the pipeline
% unchanged. \textsc{Raw} uses the mean-pooled residual-stream hidden state at
% the same layer, treating each original hidden dimension as a candidate
% utility-related feature. \textsc{PCA} uses principal components fitted on
% these hidden states, applying the same PULSE procedure in the PCA coordinate
% space.

% As shown in Table~\ref{tab:sae_need}, SAE features outperform both alternatives
% on all four datasets, improving the 4-shot average accuracy by $+3.47$ points
% over \textsc{PCA} and $+3.96$ points over \textsc{Raw}. The \textsc{Raw}
% control is particularly diagnostic: it uses the same model layer and internal
% activations as the SAE setting, but without the SAE dictionary that decomposes
% the dense residual stream into sparse coordinates. Its lower performance shows
% that the paired utility objective is less effective when applied directly to
% entangled hidden-state dimensions. Sparse SAE coordinates therefore provide a
% more effective feature space for isolating utility-related internal signals
% than either raw residual dimensions or variance-based PCA components.

\begin{wraptable}{r}{0.43\linewidth}
\vspace{-1.4em}
\centering
\small
\setlength{\tabcolsep}{3.5pt}
\renewcommand{\arraystretch}{0.92}
\caption{SAE-basis ablation on Gemma2-2B in the 4-shot pool-scale retrieval setting.}
\label{tab:sae_need}
\begin{tabular}{@{}lccc@{}}
\toprule
Dataset & SAE & PCA & Raw \\
\midrule
AGNews & \textbf{85.94} & 81.25 & 80.47 \\
REST14 & \textbf{77.73} & 75.78 & 76.76 \\
LAP14  & \textbf{77.32} & 73.22 & 74.95 \\
EMOC   & \textbf{68.36} & 65.23 & 61.33 \\
\midrule
Cls. Avg. & \textbf{77.34} & 73.87 & 73.38 \\
\midrule
CommonGen & \textbf{11.09} & 10.23 & 10.34 \\
\bottomrule
\end{tabular}
\end{wraptable}

To test whether SAE features provide a better basis for PULSE identification,
we replace them with two alternatives while keeping the paired identification
objective and retrieval pipeline unchanged: Raw uses the mean-pooled hidden
state at the same layer, and PCA uses principal components of those hidden
states. As shown in Table~\ref{tab:sae_need}, SAE features outperform both
alternatives on all four classification datasets, improving the 4-shot
classification average by $+3.47$ points over PCA and $+3.96$ points over Raw.
On CommonGen, where performance is measured by BLEU-4 rather than accuracy, SAE
also outperforms PCA and Raw, improving from $10.23$ and $10.34$ to $11.09$.
The Raw control is especially diagnostic: applied directly to entangled hidden
dimensions, the paired identification objective alone is insufficient. Sparse
SAE coordinates therefore provide a more effective feature space for isolating
utility-related internal features than either raw residual dimensions or
variance-based PCA components.
\\
\noindent\textbf{Are the PULSE-identified features useful for retrieval?}
We evaluate \textsc{PULSE-Mask}, a pure feature-masked retriever that uses
$|\mathbf{w}_l|$ without the standard SAE cosine component in
PULSE-Retriever. As shown in Table~\ref{tab:feat_valid},
\textsc{PULSE-Mask} clearly outperforms \textsc{Anti-Retrieval}, which uses
the same mask but selects the lowest-scoring pool items, improving average
accuracy from $48.52$ to $73.46$ ($+24.94$ points). This shows that the
PULSE-induced retrieval ranking is highly informative. \textsc{PULSE-Mask}
also outperforms \textsc{Bottom-Feats} and \textsc{Sparse-Random} by $+2.15$
and $+1.32$ points on average, indicating that the gain comes from the
identified utility-related features rather than sparsity alone. Finally, its
performance is comparable to \textsc{PULSE-NoTopK} ($73.46$ vs.\ $73.36$),
showing that the selected sparse feature subset retains most of the useful retrieval signal in the broader SAE feature space.

\section{Discussion}
\label{sec:discussion}

The results in Section~\ref{sec:results} show that PULSE improves both
 candidate-set scoring and pool-scale demonstration
retrieval. These gains indicate that the utility-related internal features
identified by PULSE provide a stronger selection criterion than external
query--demonstration similarity alone. We now ask two questions to better
understand these features: (1) What kinds of internal factors do high-weight
PULSE features capture? (2) To what extent is the identified feature structure
partially transferable across tasks? We first inspect representative
high-weight features, and then use cross-dataset transfer to test whether the
identified feature structure contains components that generalize beyond the
dataset on which it is estimated. The following analyses focus on the
classification setting, where feature inspection and cross-dataset transfer
can be compared under a shared evaluation protocol. Additional diagnostic
analyses, including full top-feature tables, utility-alignment visualizations,
and feature-overlap statistics, are provided in Appendix~\ref{app:feature_semantics}
and Appendix~\ref{app:cross_transfer}.

\begin{table}[t]
\centering
\small
\setlength{\tabcolsep}{4pt}
\caption{
Controls for the PULSE feature mask on Gemma2-2B in the 4-shot pool-scale retrieval setting.
\textsc{PULSE-Mask} uses only the magnitude of the PULSE-identified feature
vector, without the standard SAE cosine component used in PULSE-Retriever.
\textsc{Anti-Retrieval} uses the same mask but selects the lowest-scoring pool
items. \textsc{Sparse-Random} uses randomly selected sparse features,
\textsc{Bottom-Feats} uses features with the smallest absolute identification
scores, and \textsc{PULSE-NoTopK} removes the top-feature restriction. Accuracy
is reported in percentage points.
}
\label{tab:feat_valid}
\begin{tabular}{lccccc}
\toprule
Dataset & \textsc{PULSE-Mask} & \textsc{Anti-Retrieval} & \textsc{Sparse-Random} & \textsc{Bottom-Feats} & \textsc{PULSE-NoTopK} \\
\midrule
AGNews & \textbf{83.01} & 51.95 & 82.81 & 82.03 & 81.64 \\
REST14 & 76.17 & 63.67 & 75.39 & 75.00 & \textbf{78.12} \\
LAP14  & \textbf{74.51} & 48.38 & 72.14 & 70.41 & 74.29 \\
EMOC   & \textbf{60.16} & 30.08 & 58.20 & 57.81 & 59.38 \\
\midrule
Avg.   & \textbf{73.46} & 48.52 & 72.14 & 71.31 & 73.36 \\
\bottomrule
\end{tabular}
\vspace{-20pt}
\end{table}

\subsection{What Do High-Weight PULSE Features Capture?}
\label{sec:vis}

Because PULSE identifies utility-related features in SAE space, we can
inspect the features that receive the largest absolute weights. This analysis
does not claim that individual features causally mediate ICL performance; it
asks whether the features identified by PULSE correspond to recognizable
factors associated with demonstration utility. Table~\ref{tab:feature_evidence_cards_compact}
reports the highest-$|w|$ feature for each dataset,
together with its top-activating snippets and a post-hoc interpretation.

On AGNews, a topic classification task, the highest-weight feature activates on
technology and industry-news phrasing (e.g., ``Open Source,''
``institutional investors''), acting as a topic anchor that directly matches
the dataset's category structure. On REST14, an aspect-based sentiment task for
restaurant reviews, the selected feature captures aspect-specific evaluative
expressions such as ``desserts; got favorable reviews,'' reflecting an
aspect--opinion binding. On EMOC, an emotion recognition task over dialogues,
the feature fires on fragments containing affective reasons and topical pivots
(e.g., ``make others cry / i know whats wrong now''), both relevant cues for
emotion classification. LAP14, an aspect-based sentiment task for laptop
reviews, is especially instructive: its highest-$|w|$ feature has a negative
weight despite top-activating snippets containing positive phrases such as
``works great'' and ``power, memory and speed.'' Under the paired identification
objective, a negative weight means that higher activation is associated with
\emph{lower} utility, suggesting generic or off-aspect phrasing rather than
negative sentiment. This is consistent with PULSE-Retriever's retrieval design,
where $|\mathbf{w}_l|$ serves as a feature-relevance mask: both positive- and
negative-weight features are treated as utility-informative, differing only in
the direction of association.

Overall, high-weight PULSE features capture recognizable internal
factors---topic anchors, aspect--opinion bindings, affective cues, and
off-aspect patterns---but these examples alone do not imply that all such
features are task-specific. The next subsection tests whether this identified
feature structure also contains components that transfer across tasks.

\begin{table*}[t]
\centering
\scriptsize
\setlength{\tabcolsep}{4pt}
\renewcommand{\arraystretch}{1.12}
\caption{
Representative high-$|w|$ PULSE features. P1/N1 denotes the top positive or
negative feature after activation filtering; the sign reflects association
with paired utility differences.
}
\label{tab:feature_evidence_cards_compact}
\begin{tabularx}{\textwidth}{
@{} l c c
>{\RaggedRight\arraybackslash}p{5.8cm}
>{\RaggedRight\arraybackslash}X
@{}}
\toprule
Dataset & Feature & Rank & Top-activating snippets & Interpretation / Role \\
\midrule

AGNews
& \#8186
& P1
& \begin{tabular}[t]{@{}l@{}}
``Open Source: Balancing Innovation and Risk'' \\
``Citigroup ... institutional investors'' \\
``MFS Institutional Emerging Equities Fund''
\end{tabular}
& Institutional and technology-industry framing; task-conditioned topic anchor. \\

\specialrule{0.25pt}{0.45em}{0.45em}

REST14
& \#7875
& P1
& \begin{tabular}[t]{@{}l@{}}
``Aspect: desserts; got favorable reviews'' \\
``Aspect: service; good things to say'' \\
``Aspect: staff; good things to say''
\end{tabular}
& Aspect-specific evaluative expression; aspect--opinion binding. \\

\specialrule{0.25pt}{0.45em}{0.45em}

LAP14
& \#15777
& N1
& \begin{tabular}[t]{@{}l@{}}
``Aspect: works; Toshiba works great'' \\
``Aspect: power; power, memory and speed'' \\
``Aspect: memory; power, memory and speed''
\end{tabular}
& Generic capability or off-aspect framing; lower-utility pattern. \\

\specialrule{0.25pt}{0.45em}{0.45em}

EMOC
& \#7063
& P1
& \begin{tabular}[t]{@{}l@{}}
``make others cry / i know whats wrong now'' \\
``give love to others ... happy always'' \\
``especially yours / mine is so so / amazing''
\end{tabular}
& Self/other dialogue with affective reasoning or topical pivot; task-conditioned emotion cue. \\

\bottomrule
\end{tabularx}
\vspace{-0.5em}
\end{table*}
% ---------------------------------------------------------------
\subsection{Cross-Dataset Transfer}
\label{sec:cross_summary}

Cross-dataset transfer tests whether PULSE-identified features remain useful
beyond the dataset on which they are identified. Since each
utility-localization vector is estimated from demonstration-utility
differences on one dataset, its high-weight features may include dataset-specific information and partially transferable context-use factors.
If the vector only captured source-specific surface patterns, it should
degrade toward generic retrieval baselines when applied to another dataset.

Table~\ref{tab:cross} reports single-source transfer and a leave-one-out
variant, LOO$_\cup$ (Leave-One-Out Union): for each target dataset, this
variant excludes that target from the feature-identification stage and takes
the union of PULSE-identified features from the remaining three source
datasets, thereby testing whether utility-relevant features discovered on
other tasks can substitute for target-specific ones. The results show
the identified features are partially transferable. Averaged over all
off-diagonal source--target pairs, single-source transfer reaches $76.05$,
only $1.29$ points below the within-dataset average of $77.34$. LOO$_\cup$ is competitive with the target-trained vector: it obtains $76.71$ on
average, only $0.63$ points below Self, while outperforming the best non-PULSE
baseline by $+1.79$ points.

Transfer, however, is not uniform. LOO$_\cup$ is $2.35$ points below Self on
AGNews, suggesting that topic classification relies more on target-specific
utility information. In contrast, it matches Self on REST14, slightly exceeds
it on LAP14, and is only $0.39$ points lower on EMOC, indicating stronger
partial transfer among sentiment- and emotion-style tasks. Together with the
small cross-dataset feature overlap and mixed signs reported in
Appendix~\ref{app:cross_transfer}, these results argue against a single task-independent utility direction. Instead, PULSE identifies a mixture of
partially transferable and task-conditioned utility-related internal features.

\begin{table*}[t]
\centering
\small
\setlength{\tabcolsep}{3.2pt}
\caption{
Cross-dataset transfer of PULSE utility-localization vectors in the 4-shot
retrieval setting on Gemma2-2B. Scores are accuracy (\%).
\emph{Single-source}: a utility-localization vector learned on the source
dataset is applied directly to the target dataset.
\emph{LOO$_\cup$}: for each target dataset, the target excludes the dataset used for feature identification, and the PULSE-identified features from the three non-target
datasets are combined by union.
\emph{Base$_{\max}$}: best non-PULSE baseline per target.
$\Delta_{\mathrm{Base}}$ = LOO$_\cup$ $-$ Base$_{\max}$;
$\Delta_{\mathrm{Self}}$ = LOO$_\cup$ $-$ Self.
}
\label{tab:cross}
\begin{tabular}{l cccc ccc cc}
\toprule
& \multicolumn{4}{c}{\textbf{Single-source transfer}}
& \multicolumn{3}{c}{\textbf{LOO transfer vs. baseline}}
& \multicolumn{2}{c}{\textbf{Target PULSE}} \\
\cmidrule(lr){2-5}
\cmidrule(lr){6-8}
\cmidrule(lr){9-10}
Target
& AGNews & REST14 & LAP14 & EMOC
& LOO$_\cup$
& Base$_{\max}$
& $\Delta_{\mathrm{Base}}$
& Self
& $\Delta_{\mathrm{Self}}$ \\
\midrule
AGNews
& --    & 84.77 & 83.59 & 85.55
& 83.59 & 81.25 & $+2.34$
& 85.94 & $-2.35$ \\
REST14
& 76.56 & --    & 79.69 & 78.52
& 77.73 & 75.59 & $+2.14$
& 77.73 & \phantom{$+$}0.00 \\
LAP14
& 78.19 & 74.73 & --    & 76.46
& 77.54 & 76.03 & $+1.51$
& 77.32 & $+0.22$ \\
EMOC
& 60.94 & 66.80 & 66.80 & --
& 67.97 & 66.80 & $+1.17$
& 68.36 & $-0.39$ \\
\midrule
Avg.
& 71.90 & 75.43 & 76.69 & 80.18
& 76.71 & 74.92 & $+1.79$
& 77.34 & $-0.63$ \\
\bottomrule
\end{tabular}
\vspace{-10pt}
\end{table*}

\section{Conclusion}

We presented PULSE, an SAE-based framework for identifying
model-internal sparse features associated with in-context demonstration
utility. From paired comparisons of candidate demonstration sets, PULSE
estimates a sparse utility-localization vector over SAE features.
The PULSE-Ranking experiment validates that these identified features encode
internal utility signals through controlled complete-set ranking, while
PULSE-Retriever uses them as a feature-relevance mask for scalable pool-scale
demonstration retrieval. Experiments across classification, generation, and
reasoning tasks show that PULSE consistently improves demonstration selection
over lexical, embedding-based, SAE-space, and diversity-aware baselines.
Ablations and analyses show that the gains come from utility-aware
sparse feature identification, and that the identified features are interpretable
and partially transferable. These results support our claim that ICL
demonstration selection benefits from explicit model-internal utility-related
features, not only external query--demonstration similarity.
Extending PULSE to multimodal ICL is a possible direction; such an extension
could also test whether the identified utility signals remain reliable under
misleading visual inputs, a robustness setting evaluated by MVI-Bench
\citep{chen2025mvi}
% We presented \textbf{PULSE}, an SAE-based framework for localizing
% model-internal sparse features associated with in-context demonstration
% utility. From paired comparisons of candidate demonstration sets, PULSE learns
% a utility-localization vector over SAE features. This vector is then used in
% two ways: PULSE-Ranking experiment  applies it to controlled set-level scoring, while
% PULSE-Retriever converts it into a feature-relevance mask for scalable pool-scale
% retrieval. Experiments across four classification datasets and two backbone
% models show that PULSE-Ranking experiment  and PULSE-Retriever consistently outperform lexical,
% embedding-based, SAE-space, and diversity-aware baselines in their respective
% settings. Additional analyses indicate that the gains come from utility-aware
% sparse feature selection rather than SAE representations alone, and that the
% localized features are interpretable, partially transferable, and
% directionally aligned with downstream retrieval quality. These findings
% support the view that demonstration selection can benefit from explicit
% model-internal utility signals, not only external query--demonstration
% similarity.

{
\small
\bibliographystyle{plain}
\bibliography{refs}
}

\newpage
% ===============================================================
\appendix
% ===============================================================
\section{Experimental Details}
\FloatBarrier
\label{ref:detail}
\subsection{Model and SAE}
Because PULSE operates in SAE feature space, we evaluate it on backbone models
with publicly available SAEs: Gemma2-2B and Llama3.1-8B. Unless otherwise
specified, experiments use both backbones; GSM8K is evaluated only with
Llama3.1-8B. Table~\ref{tab:model_sae_details} reports the exact backbone
identifiers, SAE checkpoints, layers, dictionary sizes, and decoding settings.
 
\begin{table*}[t]
\centering
\scriptsize
\setlength{\tabcolsep}{2.5pt}
\renewcommand{\arraystretch}{0.95}
\caption{\textbf{Backbone, SAE, and decoding details.}
All SAE features are taken from residual-stream SAEs at the listed layer. The
Gemma-Scope checkpoint is the layer-12 width-16k average-$L_0$-82 dictionary;
the Llama-Scope LXR-32x checkpoints use $d_{\mathrm{model}}=4096$ and
$d_{\mathrm{SAE}}=131{,}072$. For classification, downstream decisions are made
by teacher-forced label likelihood rather than free-form generation.}
\label{tab:model_sae_details}
\begin{tabularx}{0.94\textwidth}{@{}l >{\raggedright\arraybackslash}p{0.25\textwidth} c c >{\raggedright\arraybackslash}X@{}}
\toprule
Backbone
& SAE release/checkpoint
& Layer(s)
& $d_{\mathrm{SAE}}$
& Inference and decoding settings \\
\midrule
Gemma2-2B
&
\path{gemma-scope-2b-pt-res-canonical}

\path{layer_12/width_16k/average_l0_82}
&
12
&
16{,}384
&
Classification: teacher-forced label likelihood, no sampling.

CommonGen: greedy generation, \texttt{do\_sample=False},
\texttt{max\_new\_tokens}=32; temperature/top-$p$ disabled. \\
\midrule
Llama3.1-8B
&
\path{llama_scope_lxr_32x}

\path{fnlp/Llama3_1-8B-Base-LXR-32x}~\cite{he2024llama}
&
\makecell[c]{12 (classification)\\
16 (CommonGen)\\
24 (GSM8K)}
&
131{,}072
&
Classification: teacher-forced label likelihood, no sampling.

CommonGen: greedy generation, \texttt{do\_sample=False},
\texttt{max\_new\_tokens}=32; temperature/top-$p$ disabled.

GSM8K: greedy generation, \texttt{do\_sample=False},
\texttt{max\_new\_tokens}=220; temperature/top-$p$ disabled. \\
\bottomrule
\end{tabularx}
\end{table*}
\subsection{Datasets}
\label{app:dataset}
We evaluate on six benchmarks spanning classification, generation, and reasoning.
For classification, we use AGNews (4-class topic classification), REST14 and
LAP14 (3-class aspect-based sentiment), and EMOC (4-class emotion recognition),
reporting classification accuracy. To keep the evaluation-query budget
comparable across classification datasets, AGNews, REST14, and EMOC are
evaluated on fixed 512-query held-out evaluation subsets, while LAP14 uses its
full 463-query test split because its test set is already of comparable size.
The same evaluation queries are used for all methods, shots, and backbones,
making all method comparisons paired within each dataset. For generation, we
use CommonGen~\cite{lin2020commongen}, reporting BLEU-4. For reasoning, we use
GSM8K~\cite{cobbe2021training}, reporting exact match accuracy. Across all
tasks, utility features are identified only from the training split, and
held-out evaluation queries are used only for downstream ICL evaluation.
Table~\ref{tab:dataset_details} summarizes the split sizes, evaluation-query
counts, input lengths, label distributions, and support-pool sizes.
 
\begin{table*}[t]
\centering
\scriptsize
\setlength{\tabcolsep}{2.5pt}
\renewcommand{\arraystretch}{1.04}
\caption{\textbf{Dataset and pool statistics.}
Train/test sizes and support-pool sizes are read from the experiment loaders.
Average input length is computed on the input side using whitespace tokenization
(concept-list length for CommonGen). Generation and reasoning experiments use
smaller pools because each selected context requires autoregressive evaluation.}
\label{tab:dataset_details}
\begin{tabularx}{0.96\textwidth}{@{}llrrrr>{\raggedright\arraybackslash}X@{}}
\toprule
Dataset & Task & Train & Test & Avg.\ len. & Pool & Label distribution \\
\midrule

AGNews & Topic cls. & 120{,}000 & 7{,}600 & 37.8 & 2{,}000 &
\makecell[l]{train: Business=30{,}000, Sci/Tech=30{,}000\\
train: Sports=30{,}000, World=30{,}000\\
test: Business=1{,}900, Sci/Tech=1{,}900\\
test: Sports=1{,}900, World=1{,}900} \\

\specialrule{0.18pt}{0.28em}{0.28em}

REST14 & Aspect sentiment & 2{,}881 & 721 & 21.9 & 2{,}000 &
\makecell[l]{train: Negative=633, Neutral=513, Positive=1{,}735\\
test: Negative=172, Neutral=120, Positive=429} \\

\specialrule{0.18pt}{0.28em}{0.28em}

LAP14 & Aspect sentiment & 1{,}850 & 463 & 23.8 & 1{,}850 &
\makecell[l]{train: Negative=686, Neutral=384, Positive=780\\
test: Negative=180, Neutral=76, Positive=207} \\

\specialrule{0.18pt}{0.28em}{0.28em}

EMOC & Emotion cls. & 38{,}424 & 5{,}509 & 17.6 & 2{,}000 &
\makecell[l]{train: angry=5{,}954, happy=4{,}669\\
train: others=21{,}963, sad=5{,}838\\
test: others=4{,}677, happy=298\\
test: angry=284, sad=250} \\

\specialrule{0.18pt}{0.28em}{0.28em}

CommonGen & Constrained gen. & 67{,}389 & 4{,}018 & 3.3 & 2{,}000 & -- \\

\specialrule{0.18pt}{0.28em}{0.28em}

GSM8K & Math reasoning & 7{,}473 & 1{,}319 & 45.3 & 2{,}000 & -- \\

\bottomrule
\end{tabularx}
\vspace{-0.6em}
\end{table*}

\subsection{Hyperparameters}
\noindent\textbf{Hyperparameter summary.}
Table~\ref{tab:hyperparams_summary} centralizes the default values for all PULSE hyperparameters used in the main experiments.
 
\begin{table*}[t]
\centering
\small
\setlength{\tabcolsep}{6pt}
\caption{\textbf{Hyperparameter summary.}
Defaults are the values used in the main experiments unless a task-specific
exception is shown.}
\label{tab:hyperparams_summary}
\begin{tabular}{lll}
\toprule
Symbol & Meaning & Default value \\
\midrule
$K^{+}$ & Positive-weight SAE features retained &
\makecell[l]{Gemma2-2B: 512\\
Llama3.1-8B: 2{,}048} \\
$K^{-}$ & Negative-weight SAE features retained &
\makecell[l]{Gemma2-2B: 512\\
Llama3.1-8B: 2{,}048} \\
$\beta$ & Sparse--dense blend weight in the retrieval score &
0.3 \\

$L$ & Retrieval shortlist size & 50 \\
$\lambda_r$ & SAE-space redundancy penalty & 0.3 \\
$|\mathcal{D}_{\mathrm{disc}}|$ & Discovery queries & 64 \\
$N$ & Candidate sets per discovery query &
\makecell[l]{Classification: 32\\
CommonGen/GSM8K: 32} \\
$\varepsilon$ & Variance-normalization stabilizer & $10^{-6}$ \\
$seed$& random seed &42\\
\bottomrule
\end{tabular}
\end{table*}

\subsection{Baseline Protocol and Tuning}
\label{app:baseline_protocol}

Table~\ref{tab:method_resources} summarizes the supervision and tuning used by
each method. PULSE uses a small labeled discovery set to estimate
$\mathbf{w}_l$, while Random, Lex-Sim, SBERT, KNN-SAE, and CEIL do not use
labeled discovery queries to learn task-specific feature weights. All methods
are evaluated on the same held-out evaluation queries, and all hyperparameters
are fixed before test evaluation.

For pool-scale retrieval, all pointwise retrieval methods use the same
shortlist-and-greedy context composition procedure with the same $L$ and
$\lambda_r$; only the relevance score used to rank candidates differs. This
ensures that differences between SBERT, KNN-SAE, and PULSE-Retriever are due to
the scoring signal rather than to a different context-composition algorithm.

For the PULSE-Ranking experiment, all methods rank the same pre-sampled
complete $k$-shot candidate sets for each query. Pointwise baselines score a
complete set by averaging query--example similarities:
\[
s_{\mathrm{base}}(q_x,E)
=
\frac{1}{|E|}
\sum_{(x_i,z_i)\in E}
\operatorname{sim}(q_x,x_i).
\]
Because all candidate sets have the same shot count, averaging and summing
produce the same ranking. CEIL uses its set-level diversity-aware objective to
score the same pre-sampled candidate sets.
\begin{table}[t]
\centering
\scriptsize
\setlength{\tabcolsep}{3pt}
\caption{
Supervision and tuning resources used by each retrieval method. PULSE uses a
small labeled discovery split to estimate a utility-localization vector; the
non-PULSE retrievers do not learn task-specific utility weights from labeled
discovery queries. For pool-scale retrieval, pointwise methods use the same
shortlist and greedy redundancy procedure unless otherwise stated.
}
\label{tab:method_resources}
\begin{tabularx}{\linewidth}{lCCCC}
\toprule
Method
& Labeled discovery data
& Validation tuning
& Model-internal SAE features
& Greedy redundancy \\
\midrule
Random & No & No & No & No \\
Lex-Sim & No & Yes & No & Yes \\
SBERT & No & Yes & No & Yes \\
KNN-SAE & No & Yes & Yes & Yes \\
CEIL & No & Yes & No & Built-in diversity objective \\
PULSE-Retriever & Yes, 64 queries & Yes & Yes & Yes \\
\bottomrule
\end{tabularx}
\end{table}
\section{Interpreting PULSE-Retriever as a Scalable Proxy}

\label{app:retriever_proxy}

\subsection{From Complete-Set Scoring to Per-Example Retrieval}
\label{app:proxy_derivation}

PULSE-Ranking uses the signed utility-localization vector at the same scale at which it is estimated. For a complete demonstration set \(E\), the ranking score is
\[
s_{\rm rank}(q_x,E)=w_l^\top\left[A_l(q_x,E)-A_l(q_x,\emptyset)\right].
\]
This score directly measures whether the complete context activates SAE features whose paired activation shifts were associated with higher or lower discovery-time utility. Applying it to all \(k\)-shot subsets of a large support pool is infeasible because the search space grows combinatorially.

A more faithful per-candidate alternative would be to score each labeled singleton demonstration by a full prompt forward pass,
\[
s_{\rm single}(q_x,x_i)=w_l^\top\left[A_l(q_x,\{(x_i,z_i)\})-A_l(q_x,\emptyset)\right].
\]
However, this would require a separate full-prompt forward pass for every query-candidate pair. Selecting the top-\(k\) singleton scores would also approximate set-level utility as an additive sum of individual utilities, ignoring interactions among examples in the final \(k\)-shot context.

PULSE-Retriever therefore uses a weaker but scalable proxy. Instead of using the sign of \(w_l\) as an individual-candidate preference direction, it uses the magnitude \(|w_l|\) as a feature-relevance mask. Let
\[
a^{(q)}=A_l(q_x,\emptyset),\qquad a^{(i)}=A_l(q_{x_i},\emptyset)
\]
be zero-shot SAE encodings of the evaluation query and candidate input, respectively, with the target omitted. The PULSE-masked similarity is
\[
s_{\rm mask}(q_x,x_i)=\cos\left(|w_l|\odot a^{(q)}, |w_l|\odot a^{(i)}\right).
\]
Equivalently, this is cosine similarity under the diagonal metric
\[
D_w=\operatorname{diag}(|w_l|^2),
\]
whose numerator is \((a^{(q)})^\top D_w a^{(i)}\). Thus, PULSE-Retriever compares the query and candidate along SAE dimensions that the discovery procedure identified as utility-related.

This proxy differs from the complete-set score in two ways. First, it uses an input-only, zero-shot representation \(A_l(q_{x_i},\emptyset)\) of the candidate rather than the activation shift produced by inserting the labeled demonstration into the query prompt. This makes candidate encodings cacheable and avoids a query-candidate full-prompt forward pass. Second, it uses \(|w_l|\) rather than signed \(w_l\), because a sign learned from set-level contrasts is not calibrated as a per-candidate preference direction. PULSE-Retriever should therefore be interpreted as a utility-informed retrieval metric, not as an exact decomposition of complete-context utility.

\subsection{Why the Retriever Uses a Magnitude Mask}
\label{app:magnitude_mask}

The distinction between PULSE-Ranking and PULSE-Retriever is a scale
distinction. At the complete-set scale, the signed vector $\mathbf{w}_l$ is
meaningful because Eq.~\ref{eq:set_score} evaluates the activation shift of the
whole context, which matches the discovery-time contrast used to estimate
$\mathbf{w}_l$. At the per-example retrieval scale, however, the retriever only
has access to input-only zero-shot encodings of the query and candidate before
the labeled demonstration is inserted into the final context. Therefore, the
set-level sign of $\mathbf{w}_l$ is not calibrated as a one-sided
individual-candidate preference direction.

PULSE-Retriever uses $|\mathbf{w}_l|$ as a feature-relevance mask:
\[
s_{\mathrm{mask}}(q_x,x_i)
=
\cos\!\bigl(
|\mathbf{w}_l|\odot\mathbf{a}^{(q)},
|\mathbf{w}_l|\odot\mathbf{a}^{(i)}
\bigr).
\]
This should be interpreted as cosine similarity under a diagonal relevance
metric over utility-informative SAE dimensions, not as a signed utility score.

A symmetric signed-mask cosine is not a distinct alternative:
\[
\cos(\mathbf{w}_l\odot\mathbf{a}^{(q)},\mathbf{w}_l\odot\mathbf{a}^{(i)})
=
\cos(|\mathbf{w}_l|\odot\mathbf{a}^{(q)},|\mathbf{w}_l|\odot\mathbf{a}^{(i)}),
\]
because the sign disappears from both the numerator and denominator. To test a
genuinely sign-sensitive retrieval rule, we therefore compare against a signed
bilinear score,
\[
s_{\mathrm{signed}}(q_x,x_i)
=
(\mathbf{a}^{(q)})^\top
\operatorname{diag}(\mathbf{w}_l)
\mathbf{a}^{(i)}.
\]
The signed score is normalized per query and uses the same shortlist and
greedy redundancy procedure as PULSE-Retriever.

Table~\ref{tab:signed_vs_abs} compares the magnitude mask with this
sign-sensitive alternative in the 4-shot pool-scale retrieval setting. The
magnitude mask performs better on average, supporting our interpretation that
the sign of $\mathbf{w}_l$ is reliable for complete-set activation shifts but
not generally calibrated as a per-example retrieval preference direction.
\begin{table}[t]
\centering
\small
\setlength{\tabcolsep}{4pt}
\caption{
Empirical comparison of the magnitude mask and a sign-sensitive retrieval
score in PULSE-Retriever. Results are from the 4-shot pool-scale retrieval
setting. The metric is accuracy for classification, BLEU-4 for CommonGen, and
exact match for GSM8K. $\Delta = |\mathbf{w}|$ mask $-$ signed score. Bold marks
$|\Delta|\ge 0.8$. ``--'' indicates GSM8K is not evaluated on Gemma2-2B.
}

\begin{tabular}{llcccccc}
\toprule
Backbone & Variant & AGNews & REST14 & LAP14 & EMOC & CommonGen & GSM8K \\
\midrule
\multirow{3}{*}{Gemma2-2B}
 & $|\mathbf{w}|$ mask & 85.94 & 77.73 & 77.32 & 68.36 & 11.09 & -- \\
 & signed score        & 84.38 & 76.56 & 77.34 & 65.23 & 10.25 & -- \\
 & $\Delta$            & \textbf{+1.56} & \textbf{+1.17} & $-0.02$ & \textbf{+3.13} & \textbf{+0.84} & -- \\
\midrule
\multirow{3}{*}{Llama3.1-8B}
 & $|\mathbf{w}|$ mask & 88.09 & 81.84 & 82.07 & 71.09 & 10.90 & 54.66 \\
 & signed score        & 85.16 & 82.03 & 82.03 & 72.66 & 10.64 & 51.53 \\
 & $\Delta$            & \textbf{+2.93} & $-0.19$ & $+0.04$ & \textbf{$-1.57$} & $+0.26$ & \textbf{+3.13} \\

\bottomrule
\end{tabular}

\label{tab:signed_vs_abs}
\end{table}
\section{Greedy Demonstration Selection}
\label{app:greedy}

Given the blended score $s_{\mathrm{blend}}$ from
Eq.~\ref{eq:blend}, each pointwise retrieval method first ranks candidate
support demonstrations and keeps a fixed-size shortlist of size $L$. We then
greedily compose the final $k$-shot demonstration context with an SAE-space
redundancy penalty.

Let $\mathrm{Sel}_{t-1}$ be the set of demonstrations selected before step
$t$. For a candidate $i$, define
\[
R_t(i)=
\begin{cases}
0, & t=1,\\[3pt]
\max\limits_{j\in \mathrm{Sel}_{t-1}}
\cos\!\bigl(\mathbf{a}^{(i)},\mathbf{a}^{(j)}\bigr), & t>1.
\end{cases}
\]
At step $t$, the next demonstration is selected as
\begin{equation}
\label{eq:greedy}
i_t
=
\arg\max_{i \in \mathrm{Shortlist}\setminus \mathrm{Sel}_{t-1}}
\left[
s_{\mathrm{blend}}(q_x,x_i)
-
\lambda_r R_t(i)
\right].
\end{equation}
The redundancy term is cosine similarity in SAE feature space, and
$\lambda_r\ge 0$ controls the penalty strength. The base case
$R_1(i)=0$ means that the first selected example is the highest-scoring
candidate in the shortlist. The exclusion
$i\in \mathrm{Shortlist}\setminus \mathrm{Sel}_{t-1}$ prevents selecting the
same demonstration more than once.
\section{Feature Semantics and Visualization Details}
\label{app:feature_semantics}

This appendix provides additional details for the feature-semantic analysis in
Section~\ref{sec:vis}. The goal of this analysis is to understand what the
high-weight coordinates in the PULSE utility-localization vector tend to represent. We
describe these features as \emph{associated with} or \emph{aligned with}
discovery-time utility, rather than as causal mediators of downstream
accuracy.

\subsection{Visualization Protocol}
\label{app:vis_protocol}

All feature visualizations in Section~\ref{sec:vis} are conducted on
Gemma2-2B with the Gemma-Scope residual-stream SAE at layer 12, matching the
main Gemma retrieval experiments. PULSE weights are learned from the training
split using 64 discovery queries. For each discovery query, candidate discovery sets are sampled
from the training support pool, and the test
split is never used during discovery-time feature identification.

For each prompt, we encode the layer-12 residual-stream activations with the
SAE and mean-pool SAE feature activations over all non-BOS prompt tokens. The
PULSE utility-localization vector is then computed from paired utility differences and SAE
activation differences, as described in Section~\ref{sec:method}. For
visualization, we search over training-pool examples and report, for each
feature, the example with the largest mean SAE activation. These snippets are
therefore illustrative max-activating examples, not manually selected
interpretations.

We apply an activation filter when constructing the displayed feature tables:
a feature is included only if its maximum activation in the train search pool
is at least $0.5$. This avoids showing high-weight features that receive large
PULSE weights during discovery but do not fire clearly on the broader
visualization corpus. Such features may still affect retrieval through small
activation changes, but they are less suitable for snippet-based semantic
interpretation.

\subsection{Representative Features Across Datasets}
\label{app:representative_features}

Tables~\ref{tab:agnews_top_features}, \ref{tab:emoc_top_features},
\ref{tab:lap14_top_features}, and \ref{tab:rest14_top_features}, report the top
five positive-weight and top five negative-weight PULSE-identified features
for each classification dataset after the activation filter. These tables
expand the qualitative summary in Section~\ref{sec:vis} by showing
max-activating snippets for the identified features. The interpretation column
is post-hoc and describes the visible pattern in the corresponding snippet.

Across the inspected high-weight features, AGNews features often correspond to
topic-anchoring cues, such as technology, geopolitical, business, or
sports-related phrasing. EMOC features often activate on dialogue fragments
with explicit affective reasons, conversational pivots, or stylistically
distinct emotional expressions. For REST14 and LAP14, several high-weight
features capture aspect--opinion bindings, while others activate on generic,
off-aspect, or decontextualized expressions.

Importantly, the sign of a feature reflects its association with paired
utility differences under the PULSE identification objective; it should not be
interpreted as sentiment polarity, label polarity, or a causal effect. Negative
weights therefore indicate features whose higher activation is associated with
lower discovery-time utility, not necessarily negative task labels.

% \begin{table}[t]
% \centering
% \small

% \label{tab:pceu_feature_cases}
% \begin{tabular}{l c c p{4.6cm} p{4.6cm}}
% \toprule
% Dataset & Feature & Weight & Max-activating snippets & Interpretation \\
% \midrule
% AGNews & \#4297  & $+4.35$ & ``Bill of Rights'', ``spam conviction'', ``Halo 2'' & digital/legal/technology news cues \\
% AGNews & \#12684 & $-3.63$ & ``Tokyo Stocks'', ``Retail sales'', ``Dodgers acquire'' & generic event or market-update style cues \\
% REST14 & \#7875  & $+0.61$ & ``desserts \ldots favorable'', ``good things \ldots service'', ``great staff'' & positive aspect--sentiment expressions \\
% REST14 & \#9781  & $-0.57$ & ``food and ambience'', ``seltzer with lime'', ``filet'' & aspect-focused food mentions with weaker sentiment signal \\
% LAP14  & \#13435 & $+1.61$ & ``wonderful price'', ``satisfied \ldots product'', ``great graphics'' & positive product-value and quality cues \\
% LAP14  & \#15777 & $-3.82$ & ``works great'', ``power, memory and speed'' & generic laptop capability cues; sign is not sentiment polarity \\
% EMOC   & \#9565  & $+10.79$ & ``love failure'', ``another guy'', ``:('' & sadness / relationship-affect cues \\
% EMOC   & \#7657  & $-8.57$  & ``real slim shady'', ``busy'', repetitive small talk & generic or repetitive conversational pattern \\
% \bottomrule
% \end{tabular}
% \caption{Representative PULSE-identified features and their max-activating examples.
% Positive and negative signs indicate association with demonstration utility.}
% \end{table}

\begin{table}[t]
\centering
\small
\caption{Top PULSE-identified features on AGNews.}
\label{tab:agnews_top_features}
\begin{tabular}{c c c p{5.0cm} p{4.0cm}}
\toprule
Sign & Feature & Weight & Max-activating snippet & Interpretation \\
\midrule
$+$ & 8186  & 7.48 & Open Source: Balancing Innovation and Risk & tech-industry commentary \\
$+$ & 11980 & 6.16 & Climate fears on sharp CO$_2$ rise & scientific finding / climate framing \\
$+$ & 14321 & 5.75 & Women Voters Enter San Diego Mayoral Fray & civic / institutional reference \\
$+$ & 4297  & 4.35 & British-Irish national kidnapped in Afghanistan & named geopolitical event \\
$+$ & 1398  & 3.70 & Australia struggling against New Zealand & sports-news event \\
\midrule
$-$ & 12684 & -3.65 & Bond Prices Edge Higher at Close of Day & market-summary phrasing \\
$-$ & 12664 & -3.52 & Keep low profile, US citizens urged & security-advisory boilerplate \\
$-$ & 7429  & -2.38 & Owens' comments weren't necessary & sports-quote commentary \\
$-$ & 8762  & -2.32 & Dying star creates fantasy-like sculpture & descriptive scientific narrative \\
$-$ & 13690 & -2.24 & IRL: La der pour Castroneves & non-English race recap \\
\bottomrule
\end{tabular}
\end{table}

\begin{table}[t]
\centering
\small
\caption{Top PULSE-identified features on EMOC.}
\label{tab:emoc_top_features}
\begin{tabular}{c c c p{5.0cm} p{4.0cm}}
\toprule
Sign & Feature & Weight & Max-activating snippet & Interpretation \\
\midrule
$+$ & 7063  & 11.19 & no / make others cry / i know whats wrong now & dialogue with topical pivot \\
$+$ & 15322 & 11.18 & shut up / my exam is on tuesday & direct-address with factual follow-up \\
$+$ & 9565  & 10.81 & you are very bad / sorry / because\ldots & confrontation with stated reason \\
$+$ & 5466  & 10.68 & what else do you know about me? & introspective Q--A small talk \\
$+$ & 15273 & 9.54  & what happened / net problems on my phone & factual problem report and topic shift \\
\midrule
$-$ & 7657  & -8.59 & me happy / happy happy warm warm & repetitive affect tokens \\
$-$ & 5671  & -8.40 & turtles are stupid / i am also a turtle & absurdist syllogism \\
$-$ & 5435  & -8.38 & feeling good, feeling great !!! & repeated affect words and emoji \\
$-$ & 12481 & -8.33 & nice joke / mom you are perfect & joke-request loop \\
$-$ & 3537  & -8.32 & i don't like much / neither do i & flat agreement / topic decline \\
\bottomrule
\end{tabular}
\end{table}

\begin{table}[t]
\centering
\small
\caption{Top PULSE-identified features on LAP14.}
\label{tab:lap14_top_features}
\begin{tabular}{c c c p{5.0cm} p{4.0cm}}
\toprule
Sign & Feature & Weight & Max-activating snippet & Interpretation \\
\midrule
$+$ & 11485 & 3.13 & Aspect: screen; big screen allows you to enjoy movies & aspect to enjoyment \\
$+$ & 5828  & 1.68 & Aspect: Pages; design tools make it easier & software aspect to ease of use \\
$+$ & 13435 & 1.61 & Aspect: PRICE; LOVE THIS LAPTOP WONDERFUL PRICE & aspect to strong positive judgment \\
$+$ & 16061 & 1.56 & Aspect: battery; why not give you a better battery? & aspect to complaint \\
$+$ & 9749  & 1.46 & Aspect: price; iWork is not worth the price & aspect to negative judgment \\
\midrule
$-$ & 15777 & -3.82 & Aspect: works; my new Toshiba works great on both & generic verb / off-aspect framing \\
$-$ & 15339 & -3.11 & same Toshiba sentence & duplicate off-aspect framing \\
$-$ & 517   & -2.94 & Aspect: Powerpoint program; PC users use Powerpoint & descriptive use-case \\
$-$ & 16208 & -2.12 & Aspect: zooming; Wonderful zooming & decontextualized short judgment \\
$-$ & 11584 & -2.09 & Aspect: wireless mouse; an absolute must & necessity claim without opinion detail \\
\bottomrule
\end{tabular}
\end{table}

\begin{table}[t]
\centering
\small
\caption{Top PULSE-identified features on REST14.}
\label{tab:rest14_top_features}
\begin{tabular}{c c c p{5.0cm} p{4.0cm}}
\toprule
Sign & Feature & Weight & Max-activating snippet & Interpretation \\
\midrule
$+$ & 7875 & 0.61 & Aspect: desserts; got favorable reviews & committed positive judgment \\
$+$ & 5435 & 0.56 & Aspect: owner; always has a smile & warm person-description \\
$+$ & 1512 & 0.53 & Aspect: food; good food and ok service & committed mixed judgment \\
$+$ & 3486 & 0.44 & Aspect: Decor; leaves something to be desired & committed negative judgment \\
$+$ & 8374 & 0.43 & Aspect: comfort food; hot and large portions & simple positive description \\
\midrule
$-$ & 9781  & -0.57 & Aspect: ambience; I care more about the food and ambience & meta-commentary \\
$-$ & 13165 & -0.56 & Aspect: sour spicy soup; drumsticks over rice and soup & enumerative menu list \\
$-$ & 9472  & -0.54 & Aspect: thai; definitely not great & generic-aspect reference \\
$-$ & 11584 & -0.54 & Aspect: privacy; corner booth & parenthetical aside / off-aspect framing \\
$-$ & 12741 & -0.52 & Aspect: meal; amazing meal and experience & generic aspect with strong sentiment \\
\bottomrule
\end{tabular}
\end{table}

\section{Cross-Dataset Transfer}
\label{app:cross_transfer}

This appendix expands on the cross-dataset transfer summary in
Section~\ref{sec:cross_summary}. We evaluate three questions: whether PULSE
weights learned on one dataset transfer to another, how much active feature
support is shared across datasets, and whether shared features are used in
the same direction.

\subsection{Source--Target Transfer}
\label{app:transfer_matrix}

We learn the PULSE utility-localization vector on a source dataset and apply it directly to
pool-scale retrieval on a different target dataset, in the 4-shot retriever
setting with $\beta=0.3$. We additionally evaluate LOO$_\cup$, which merges
the vector learned from the three non-target datasets.

Table~\ref{tab:cross} shows substantial cross-task reuse. Averaged over all
off-diagonal source--target pairs, transfer accuracy is $76.05$, only
$1.29$ points below the within-dataset \textsc{Self} average of $77.34$.
LOO$_\cup$ matches or nearly matches \textsc{Self} on three targets: it
equals \textsc{Self} on REST14, slightly exceeds it on LAP14, and is only
$0.39$ points lower on EMOC. The larger gap on AGNews suggests that topic
classification benefits more from task-specific utility features.

\subsection{Feature Overlap and Sign Consistency}
\label{app:overlap_sign}

To understand how transfer is possible despite task-specific weights, we
examine the support and signs of active PULSE features across datasets.
Table~\ref{tab:feature_reuse} shows that active-feature overlap is generally
small. The largest overlap appears between REST14 and LAP14, which share
$207$ active features with a Jaccard overlap of $0.112$, consistent with
their shared aspect-based sentiment structure. All other dataset pairs have
Jaccard overlaps below $0.075$.

\begin{table}[t]
\centering
\small
\setlength{\tabcolsep}{5pt}
\caption{Pairwise overlap of active PULSE features across datasets. The two
aspect-based sentiment datasets, REST14 and LAP14, have the largest overlap,
but the overall Jaccard values remain small.}
\label{tab:feature_reuse}
\begin{tabular}{lccc}
\toprule
Dataset pair & Intersection & Union & Jaccard \\
\midrule
AGNews--EMOC   & 94  & 1954 & 0.048 \\
AGNews--REST14 & 90  & 1958 & 0.046 \\
AGNews--LAP14  & 87  & 1961 & 0.044 \\
EMOC--REST14   & 140 & 1908 & 0.073 \\
EMOC--LAP14    & 95  & 1953 & 0.049 \\
REST14--LAP14  & 207 & 1841 & 0.112 \\
\bottomrule
\end{tabular}
\end{table}

Sign consistency further shows that shared features are not used in the same
way across tasks. As shown in Table~\ref{tab:sign_mix}, $44.1\%$ of features
active in exactly two datasets have mixed signs; this ratio rises to
$60.9\%$ for features active in three datasets and $75.0\%$ for features
active in all four. Thus, the same SAE coordinate can be utility-positive for
one task but utility-negative for another.

\begin{table}[t]
\centering
\small
\setlength{\tabcolsep}{5pt}
\caption{Sign consistency among active PULSE features across datasets.
A mixed-sign feature is utility-positive on at least one dataset and
utility-negative on at least one other dataset.}
\label{tab:sign_mix}
\begin{tabular}{cccccc}
\toprule
\# active datasets & \# features & Consistent $+$ & Consistent $-$ & Mixed & \% mixed \\
\midrule
1 & 2894 & 1401 & 1493 & --  & -- \\
2 & 497  & 145  & 133  & 219 & 44.1 \\
3 & 64   & 19   & 6    & 39  & 60.9 \\
4 & 4    & 1    & 0    & 3   & 75.0 \\
\bottomrule
\end{tabular}
\end{table}

Together, these results suggest that PULSE uses reusable SAE coordinates but
assigns task-dependent signs and magnitudes. The learned signal is therefore
partly transferable, but not a single task-independent utility direction.

\section{Learned Retriever Baseline and Feedback Budget}
\label{app:epr_budget}
 
The main experiments compare PULSE-Retriever against unsupervised
baselines that do not consume target-model feedback. To test whether the
gains stem from the SAE-based identification mechanism rather than from
access to labeled feedback alone, we additionally compare against an
EPR-style learned retriever \citep{rubin2022learning}, which also uses
target-model feedback but channels it into gradient-based retriever
training.
 
The EPR-style baseline retrieves BM25 top-$16$ candidates for each of
$128$ training queries, evaluates each singleton query--demonstration pair
with target-model feedback, and trains a dense retriever head on the
resulting $2{,}048$ labeled pairs. PULSE-Retriever uses only $64$
discovery queries, requires no gradient-based training, and produces a
closed-form utility-localization vector over SAE features.
 
\begin{table*}[t]
\centering
\scriptsize
\setlength{\tabcolsep}{4pt}
\caption{
Comparison with a supervised learned retriever in the 4-shot pool-scale
retrieval setting. Classification metrics are accuracy (\%);
CommonGen is BLEU-4; GSM8K is exact match (\%). Cls.\ Avg.\ is the
mean over four classification datasets. The learned retriever uses 128
training queries expanded into $2{,}048$ singleton feedback labels and
gradient-based retriever training; PULSE-Retriever uses 64 discovery
queries with closed-form feature scoring and no retriever training.
}
\resizebox{\textwidth}{!}{
\begin{tabular}{llccccc|c|cc}
\toprule
 & & \multicolumn{4}{c}{Classification (Acc.)} & & & \multicolumn{2}{c}{Generation \& Reasoning} \\
\cmidrule(lr){3-6} \cmidrule(lr){9-10}
Backbone & Method
& AGNews & REST14 & LAP14 & EMOC & Cls.\ Avg.
& Query budget
& CommonGen & GSM8K \\
\midrule
\multirow{2}{*}{Gemma2-2B}
& EPR-style learned retriever
& \textbf{87.50} & 75.00 & \textbf{80.35} & 66.21 & 77.27
& 128 queries
& 9.78 & -- \\
& PULSE-Retriever
& 85.94 & \textbf{77.73} & 77.32 & \textbf{68.36} & \textbf{77.34}
& 64 queries
& \textbf{11.09} & -- \\
\midrule
\multirow{2}{*}{Llama3.1-8B}
& EPR-style learned retriever
& \textbf{89.26} & 77.73 & \textbf{83.00} & 70.89 & 80.22
& 128 queries
& 10.19 & 44.53 \\
& PULSE-Retriever
& 88.09 & \textbf{81.84} & 82.07 & \textbf{71.09} & \textbf{80.77}
& 64 queries
& \textbf{10.90} & \textbf{54.66} \\
\bottomrule
\end{tabular}
}

\label{tab:epr_budget}
\end{table*}
 
As shown in Table~\ref{tab:epr_budget}, PULSE-Retriever achieves a
higher classification average than the learned retriever on both
backbones (77.34 vs.\ 77.27 on Gemma2-2B; 80.77 vs.\ 80.22 on
Llama3.1-8B) despite using half the query budget and no gradient-based
retriever training. The two methods show complementary per-dataset
strengths on classification: the learned retriever leads on AGNews and
LAP14, while PULSE-Retriever leads on REST14 and EMOC across both
backbones. On generation and reasoning tasks, PULSE-Retriever shows a
clearer advantage: it outperforms the learned retriever on CommonGen by
+1.31 BLEU-4 on Gemma2-2B and +0.71 on Llama3.1-8B, and by +10.13
exact match on GSM8K, where the learned retriever falls below the
random baseline (44.53 vs.\ 47.61 in Table~3).
 
These results indicate that the improvements over unsupervised baselines
in Tables~1--2 are not merely a consequence of having access to labeled
feedback; the SAE feature space provides a utility identification signal
that a standard learned retriever does not capture. The advantage is
especially pronounced on generation and reasoning tasks, where
singleton feedback labels may be insufficient to capture the set-level
utility structure that PULSE identifies through paired comparisons.
Moreover, unlike the learned retriever, PULSE-Retriever produces an
interpretable utility-localization vector that supports the feature
inspection (Section \ref{sec:vis}) and cross-task transfer (Section \ref{sec:cross_summary}) analyses
presented in this work.
\section{Hyperparameter Sensitivity}
\label{app:hyperparam}

We report additional sensitivity analyses for three implementation choices in
PULSE: the number of selected SAE features $K$, the number of discovery queries
$|\mathcal{D}_{\mathrm{disc}}|$, and the sparse--dense blend weight $\beta$.
All experiments are conducted in the 4-shot pool-scale retrieval setting.

\noindent\textbf{Number of identified features.}
Table~\ref{tab:app-feature-budget} shows the effect of varying the number of
PULSE-identified features. Performance is stable across a wide range of feature
budgets, with the average accuracy ranging from $76.62$ to $77.65$. This
suggests that PULSE does not rely on a finely tuned feature count; once enough
high-utility SAE features are retained, the retrieval signal remains stable.

\begin{table}[t]
\centering
\small
\caption{
Retriever accuracy as a function of selected feature count $K$ in the 4-shot
pool-scale retrieval setting. Bold indicates the best result in each row.
}
\label{tab:app-feature-budget}
\begin{tabular}{lccccc}
\toprule
Dataset & K=256 & K=512 & K=1024 & K=2048 & K=4096 \\
\midrule
AGNews & 85.16 & 85.94 & \textbf{86.13} & 85.94 & 84.38 \\
REST14 & 78.12 & 77.73 & 79.30 & \textbf{80.08} & 79.69 \\
LAP14  & 77.97 & 77.32 & 76.67 & 78.19 & \textbf{78.62} \\
EMOC   & 65.23 & \textbf{68.36} & 68.16 & 66.41 & 66.80 \\
\midrule
Avg.   & 76.62 & 77.34 & 77.57 & \textbf{77.65} & 77.37 \\
\bottomrule
\end{tabular}
\end{table}

\noindent\textbf{Discovery set size.}
Table~\ref{tab:app-disc-size} reports performance as the number of discovery
queries varies from $16$ to $128$. The average accuracy improves from $75.53$
with $16$ discovery queries to $77.34$ with $64$ queries, and then saturates at
$128$ queries. This indicates that PULSE can identify a useful sparse utility
vector from a relatively small labeled discovery set.

\begin{table}[t]
\centering
\small
\caption{
Retriever accuracy as a function of discovery query count
$|\mathcal{D}_{\mathrm{disc}}|$ in the 4-shot pool-scale retrieval setting.
Bold indicates the best result in each row.
}
\label{tab:app-disc-size}
\begin{tabular}{lcccc}
\toprule
Dataset & 16 & 32 & 64 & 128 \\
\midrule
AGNews & 82.42 & 83.98 & 85.94 & \textbf{86.72} \\
REST14 & 77.73 & \textbf{78.52} & 77.73 & 77.34 \\
LAP14  & \textbf{79.48} & 78.19 & 77.32 & 78.40 \\
EMOC   & 62.50 & 64.45 & \textbf{68.36} & 66.80 \\
\midrule
Avg.   & 75.53 & 76.28 & \textbf{77.34} & 77.32 \\
\bottomrule
\end{tabular}
\end{table}

\noindent\textbf{Sparse--dense blend weight.}
Table~\ref{tab:app-beta} evaluates the blend weight $\beta$ between
PULSE-masked similarity and standard SAE-space cosine similarity. Pure
PULSE weight retrieval ($\beta=0$) and pure SAE cosine retrieval ($\beta=1$)
are both weaker than intermediate blend settings on average. The best average
performance is obtained at $\beta=0.3$, suggesting that the learned
utility-related feature mask and broader SAE-space similarity provide
complementary signals.

\begin{table}[t]
\centering
\small
\setlength{\tabcolsep}{4pt}
\caption{
Retriever accuracy as a function of the sparse--dense blend weight $\beta$ in
the 4-shot pool-scale retrieval setting. $\beta=0$ denotes pure PULSE-weighted
cosine similarity, while $\beta=1$ denotes pure SAE-space cosine similarity.
Bold indicates the best result in each row.
}
\label{tab:app-beta}
\begin{tabular}{lccccccc}
\toprule
Dataset & $\beta=0.0$ & $\beta=0.1$ & $\beta=0.2$ & $\beta=0.3$ & $\beta=0.5$ & $\beta=0.7$ & $\beta=1.0$ \\
\midrule
AGNews & 83.01 & 83.20 & 85.94 & 85.94 & \textbf{86.72} & 86.52 & 81.05 \\
LAP14  & 74.51 & 75.59 & 74.30 & \textbf{77.32} & 76.03 & 75.16 & 75.16 \\
REST14 & 76.17 & 75.59 & 76.95 & 77.73 & \textbf{77.93} & 76.76 & 75.59 \\
EMOC   & 60.16 & 63.87 & 62.30 & \textbf{68.36} & 67.77 & 66.41 & 66.80 \\
\midrule
Avg.   & 73.46 & 74.56 & 74.87 & \textbf{77.34} & 77.11 & 76.21 & 74.65 \\
\bottomrule
\end{tabular}
\end{table}
\noindent\textbf{SAE layer choice.}
We further evaluate the sensitivity of PULSE-Retriever to the SAE layer on Gemma2-2B. Table~\ref{tab:app-layer-summary} summarizes the layer sweep by reporting the best layer, the best average accuracy, and the performance of the default layer used in the main experiments. Tables~\ref{tab:app-layer-1shot}--\ref{tab:app-layer-8shot} provide the complete per-layer results on AGNews, LAP14, REST14, and EMOC.

Overall, PULSE-Retriever is stable across a broad range of SAE layers. Although the best layer varies across shot settings, the default Layer~12 remains close to the best average performance, with gaps of at most 1.14 percentage points. This suggests that PULSE does not rely on a narrowly tuned layer choice.

\begin{table}[t]
\centering
\small
\setlength{\tabcolsep}{5pt}
\renewcommand{\arraystretch}{0.95}
\caption{
Summary of SAE layer sensitivity for PULSE-Retriever on Gemma2-2B in the pool-scale retrieval setting. Best Avg. is the highest average accuracy across layers. Layer-12 Avg. denotes the default layer used in the main experiments. Gap is the difference between Best Avg. and Layer-12 Avg.; Layer Range is the difference between the best and worst layer averages. All numeric entries except layer indices are reported in percentages or percentage points.
}
\label{tab:app-layer-summary}
\begin{tabular}{lccccc}
\toprule
Shot & Best Layer & Best Avg. (\%) & Layer-12 Avg. (\%) & Gap (pp) & Layer Range (pp) \\
\midrule
1-shot & 12 & \textbf{70.64} & 70.64 & 0.00 & 4.86 \\
2-shot & 21 & \textbf{75.02} & 74.31 & 0.71 & 2.57 \\
4-shot & 5 & \textbf{78.48} & 77.34 & 1.14 & 3.22 \\
8-shot & 24 & \textbf{80.10} & 79.61 & 0.49 & 2.40 \\
\bottomrule
\end{tabular}
\end{table}

\begin{table}[t]
\centering
\scriptsize
\setlength{\tabcolsep}{3.1pt}
\renewcommand{\arraystretch}{0.92}
\caption{
Full SAE layer sweep for PULSE-Retriever on Gemma2-2B in the 1-shot pool-scale
retrieval setting. Each entry reports downstream accuracy (\%). Avg. denotes the average across AGNews, LAP14, REST14, and EMOC. $^\dagger$ denotes the default layer used in the main experiments. Bold indicates the best value in each row.
}
\label{tab:app-layer-1shot}
\resizebox{\textwidth}{!}{%
\begin{tabular}{l*{13}{c}}
\toprule
Layer & 0 & 1 & 2 & 3 & 4 & 5 & 6 & 7 & 8 & 9 & 10 & 11 & 12$^\dagger$ \\
\midrule
AGNews & 73.24 & 79.49 & 77.54 & 76.37 & 74.80 & 73.63 & 75.00 & 77.54 & 73.63 & 79.49 & 78.91 & 72.66 & \textbf{81.25} \\
LAP14 & 67.39 & 65.44 & 69.55 & 67.60 & 67.39 & 70.41 & 68.68 & 68.25 & 66.95 & 68.68 & 71.92 & 69.33 & \textbf{72.79} \\
REST14 & 64.26 & 60.94 & 65.43 & 65.82 & \textbf{67.58} & 65.82 & 66.80 & 65.62 & 66.02 & 65.62 & 64.65 & 66.60 & 64.65 \\
EMOC & 62.11 & 62.50 & 63.28 & 62.70 & 65.23 & 64.06 & 60.74 & 63.28 & \textbf{65.43} & 58.40 & 61.52 & 62.30 & 63.87 \\
Avg. & 66.75 & 67.09 & 68.95 & 68.12 & 68.75 & 68.48 & 67.81 & 68.67 & 68.01 & 68.05 & 69.25 & 67.72 & \textbf{70.64} \\
\midrule
Layer & 13 & 14 & 15 & 16 & 17 & 18 & 19 & 20 & 21 & 22 & 23 & 24 & 25 \\
\midrule
AGNews & 71.68 & 76.95 & 72.27 & 71.68 & 75.00 & 77.93 & 77.54 & 79.49 & 75.20 & 75.00 & 80.86 & 74.41 & 75.98 \\
LAP14 & 68.03 & 66.74 & 67.82 & 66.95 & 69.55 & 71.27 & 70.41 & 68.90 & 68.47 & 68.90 & 68.03 & 67.60 & 66.74 \\
REST14 & 65.43 & 65.82 & 64.45 & 64.84 & 67.19 & 65.04 & \textbf{67.58} & 67.19 & 66.80 & 64.06 & 63.48 & 66.60 & 65.82 \\
EMOC & 64.65 & 57.23 & 58.59 & 59.77 & 57.62 & 64.65 & 59.96 & 62.70 & 60.74 & 64.06 & 58.79 & 59.77 & 56.45 \\
Avg. & 67.45 & 66.68 & 65.78 & 65.81 & 67.34 & 69.72 & 68.87 & 69.57 & 67.80 & 68.01 & 67.79 & 67.10 & 66.25 \\
\bottomrule
\end{tabular}%
}
\end{table}

\begin{table}[t]
\centering
\scriptsize
\setlength{\tabcolsep}{3.1pt}
\renewcommand{\arraystretch}{0.92}
\caption{
Full SAE layer sweep for PULSE-Retriever on Gemma2-2B in the 2-shot pool-scale
retrieval setting. Each entry reports downstream accuracy (\%). Avg. denotes the average across AGNews, LAP14, REST14, and EMOC. $^\dagger$ denotes the default layer used in the main experiments. Bold indicates the best value in each row.
}
\label{tab:app-layer-2shot}
\resizebox{\textwidth}{!}{%
\begin{tabular}{l*{13}{c}}
\toprule
Layer & 0 & 1 & 2 & 3 & 4 & 5 & 6 & 7 & 8 & 9 & 10 & 11 & 12$^\dagger$ \\
\midrule
AGNews & 80.86 & 76.95 & 79.49 & 79.30 & 82.23 & 80.66 & 80.86 & 83.20 & 83.98 & 81.64 & 81.84 & 83.01 & 82.62 \\
LAP14 & 75.59 & 74.30 & 74.95 & 74.51 & 74.51 & 75.16 & 72.57 & 73.65 & 73.43 & 73.43 & 76.46 & 75.38 & \textbf{77.11} \\
REST14 & 73.05 & 68.95 & 69.53 & 72.07 & 73.44 & 73.44 & 74.22 & 70.70 & 73.24 & 73.83 & 71.29 & \textbf{75.00} & 71.68 \\
EMOC & 69.14 & \textbf{72.07} & 69.34 & 70.51 & 65.82 & 68.36 & 68.95 & 65.82 & 68.36 & 67.38 & 67.77 & 66.02 & 65.82 \\
Avg. & 74.66 & 73.07 & 73.33 & 74.10 & 74.00 & 74.41 & 74.15 & 73.34 & 74.76 & 74.07 & 74.34 & 74.85 & 74.31 \\
\midrule
Layer & 13 & 14 & 15 & 16 & 17 & 18 & 19 & 20 & 21 & 22 & 23 & 24 & 25 \\
\midrule
AGNews & 83.20 & 83.20 & 82.42 & 85.55 & 84.38 & 84.57 & \textbf{86.13} & 83.01 & 84.38 & 83.79 & 85.35 & 82.62 & 84.96 \\
LAP14 & 75.81 & 74.30 & 75.38 & 74.08 & 75.81 & 74.08 & 74.51 & 73.00 & 76.24 & 74.08 & 75.81 & 76.03 & 73.00 \\
REST14 & 73.63 & 73.05 & 70.90 & 72.66 & 70.31 & 70.51 & 70.31 & 72.66 & 72.27 & 69.14 & 70.70 & 71.48 & 71.48 \\
EMOC & 66.99 & 62.70 & 67.77 & 67.19 & 66.02 & 70.51 & 64.26 & 63.67 & 67.19 & 67.19 & 63.67 & 66.41 & 60.35 \\
Avg. & 74.91 & 73.31 & 74.12 & 74.87 & 74.13 & 74.92 & 73.80 & 73.08 & \textbf{75.02} & 73.55 & 73.88 & 74.13 & 72.45 \\
\bottomrule
\end{tabular}%
}
\end{table}

\begin{table}[t]
\centering
\scriptsize
\setlength{\tabcolsep}{3.1pt}
\renewcommand{\arraystretch}{0.92}
\caption{
Full SAE layer sweep for PULSE-Retriever on Gemma2-2B in the 4-shot pool-scale
retrieval setting. Each entry reports downstream accuracy (\%). Avg. denotes the average across AGNews, LAP14, REST14, and EMOC. $^\dagger$ denotes the default layer used in the main experiments. Bold indicates the best value in each row.
}
\label{tab:app-layer-4shot}
\resizebox{\textwidth}{!}{%
\begin{tabular}{l*{13}{c}}
\toprule
Layer & 0 & 1 & 2 & 3 & 4 & 5 & 6 & 7 & 8 & 9 & 10 & 11 & 12$^\dagger$ \\
\midrule
AGNews & 84.77 & 84.18 & 83.20 & 85.55 & 83.40 & 86.52 & 86.33 & 85.16 & 86.33 & 88.67 & 84.77 & 85.55 & 85.94 \\
LAP14 & 75.59 & 77.97 & 77.32 & 75.81 & 77.54 & 77.97 & 79.70 & 77.75 & 77.75 & 77.54 & 77.54 & 77.32 & 77.32 \\
REST14 & 75.59 & 73.63 & 75.78 & 75.39 & 73.44 & \textbf{77.93} & 74.80 & 75.39 & 75.98 & 75.39 & 74.41 & 75.00 & 77.73 \\
EMOC & 70.51 & 71.29 & 69.34 & 71.09 & 71.48 & 71.48 & 67.58 & 69.53 & \textbf{72.07} & 68.95 & 69.92 & 70.12 & 68.36 \\
Avg. & 76.61 & 76.77 & 76.41 & 76.96 & 76.46 & \textbf{78.48} & 77.10 & 76.96 & 78.03 & 77.64 & 76.66 & 77.00 & 77.34 \\
\midrule
Layer & 13 & 14 & 15 & 16 & 17 & 18 & 19 & 20 & 21 & 22 & 23 & 24 & 25 \\
\midrule
AGNews & 85.94 & 86.52 & 85.16 & 86.52 & 84.96 & 86.91 & 88.09 & \textbf{89.06} & 86.13 & 86.91 & 88.87 & 87.70 & 84.57 \\
LAP14 & 78.19 & 77.97 & 79.27 & 78.83 & 77.75 & 77.32 & 78.62 & \textbf{80.13} & 76.24 & 77.54 & 77.32 & 76.89 & 78.19 \\
REST14 & 75.59 & \textbf{77.93} & 75.78 & 75.59 & 75.98 & 75.59 & 74.61 & 73.63 & 75.39 & 75.39 & 75.20 & 73.83 & 76.37 \\
EMOC & 68.95 & 66.99 & 66.99 & 69.14 & 69.73 & 70.12 & 68.75 & 66.80 & 67.97 & 65.43 & 68.36 & 66.60 & 61.91 \\
Avg. & 77.16 & 77.35 & 76.80 & 77.52 & 77.10 & 77.48 & 77.52 & 77.41 & 76.43 & 76.32 & 77.44 & 76.25 & 75.26 \\
\bottomrule
\end{tabular}%
}
\end{table}

\begin{table}[t]
\centering
\scriptsize
\setlength{\tabcolsep}{3.1pt}
\renewcommand{\arraystretch}{0.92}
\caption{
Full SAE layer sweep for PULSE-Retriever on Gemma2-2B in the 8-shot pool-scale
retrieval setting. Each entry reports downstream accuracy (\%). Avg. denotes the average across AGNews, LAP14, REST14, and EMOC. $^\dagger$ denotes the default layer used in the main experiments. Bold indicates the best value in each row.
}
\label{tab:app-layer-8shot}
\resizebox{\textwidth}{!}{%
\begin{tabular}{l*{13}{c}}
\toprule
Layer & 0 & 1 & 2 & 3 & 4 & 5 & 6 & 7 & 8 & 9 & 10 & 11 & 12$^\dagger$ \\
\midrule
AGNews & 86.52 & 85.74 & 85.55 & 83.59 & \textbf{90.04} & 86.13 & 85.55 & 88.28 & 88.67 & 88.48 & 86.91 & 87.50 & 88.09 \\
LAP14 & 78.40 & 77.97 & 78.62 & 79.91 & 79.70 & 79.27 & 79.48 & 80.35 & 79.27 & 80.78 & 79.48 & 79.48 & 80.35 \\
REST14 & 78.32 & 78.91 & 76.95 & 77.34 & 76.37 & 77.15 & 77.73 & 77.73 & 78.32 & 77.73 & 76.17 & \textbf{79.49} & 77.73 \\
EMOC & 73.83 & 73.44 & 74.80 & 73.83 & 73.05 & 72.85 & \textbf{75.00} & 73.44 & 72.85 & 70.31 & 71.68 & 70.31 & 72.27 \\
Avg. & 79.27 & 79.01 & 78.98 & 78.67 & 79.79 & 78.85 & 79.44 & 79.95 & 79.78 & 79.33 & 78.56 & 79.20 & 79.61 \\
\midrule
Layer & 13 & 14 & 15 & 16 & 17 & 18 & 19 & 20 & 21 & 22 & 23 & 24 & 25 \\
\midrule
AGNews & 87.50 & 87.30 & 88.28 & 89.06 & 87.30 & 88.87 & 88.28 & 87.89 & 88.87 & 89.26 & 89.65 & 88.48 & 88.09 \\
LAP14 & 79.05 & 80.13 & \textbf{80.99} & 79.70 & 77.97 & 78.19 & 77.97 & 80.56 & 79.27 & 79.27 & 79.48 & 80.56 & 80.13 \\
REST14 & 77.34 & 78.71 & 78.91 & 78.71 & 76.76 & 77.15 & 77.93 & 76.95 & 77.73 & 77.54 & 77.73 & 79.30 & 77.34 \\
EMOC & 69.14 & 68.95 & 70.51 & 70.90 & 70.12 & 69.73 & 70.12 & 69.92 & 68.75 & 70.70 & 71.88 & 72.07 & 65.23 \\
Avg. & 78.26 & 78.77 & 79.67 & 79.59 & 78.04 & 78.48 & 78.57 & 78.83 & 78.65 & 79.19 & 79.68 & \textbf{80.10} & 77.70 \\
\bottomrule
\end{tabular}%
}
\end{table}

\section{Compute Resources}
\label{Compute resources}
Experiments were run as local single-GPU jobs on NVIDIA GPUs. The primary
workstation used for the reported runs contains 8*NVIDIA GeForce RTX 5090
(32GB) and 8*NVIDIA GeForce RTX 3090 (24GB); each run was assigned to one GPU
at a time. We use frozen backbone language models and pretrained SAEs, without
fine-tuning the language models or training new SAEs. The compute cost is
dominated by forward passes for discovery-time paired-utility estimation and
downstream ICL evaluation. SAE activations and candidate-level intermediate
results are cached and reused across methods where possible. In our Gemma2-2B
classification logs, constructing the static SAE cache for a pool of
1.85k--2k examples took about 141--277 seconds per dataset, while the
discovery forward phase with 64 discovery queries and 32 candidate discovery sets per discovery query took about 448--884 seconds per dataset. As a hardware-independent
scale estimate, a dense forward pass costs roughly $2P$ floating-point
operations per token for a model with $P$ parameters, corresponding to
approximately 5.2 GFLOPs/token for Gemma2-2B and 16 GFLOPs/token for
Llama3.1-8B, plus the SAE hook cost. Using the observed Gemma2-2B prompt
lengths in the classification setting, this corresponds to roughly
0.25--0.6 PFLOPs for constructing one static pool cache and roughly
7--16 PFLOPs for one discovery pass. We report these as approximate ranges
rather than exact total FLOPs because the total depends on prompt length,
label-word batching, generation length, and cache reuse.

\section{Seed Stability of PULSE-Retriever}
\label{app:seed_stability}

To assess the stability of PULSE-Retriever across random seeds, we run the
4-shot pool-scale retrieval experiment with three independent seeds. Each seed
controls the sampling of the discovery set $\mathcal{D}_{\mathrm{disc}}$ from
the training split and the sampling of the $N$ candidate demonstration sets
per discovery query, while the test split and evaluation protocol are held
fixed. Table~\ref{tab:seed_stability} reports the per-seed accuracy together
with the mean and standard deviation across seeds for both backbones.

\begin{table}[t]
\centering
\small
\caption{Seed stability of PULSE-Retriever in the 4-shot pool-scale retrieval
setting. We report accuracy (\%) for classification, BLEU-4 for CommonGen,
and exact match (\%) for GSM8K. Each row reports the mean $\pm$ standard
deviation over three seeds, along with the individual seed values. GSM8K is
not evaluated on Gemma2-2B.}
\label{tab:seed_stability}
\begin{tabular}{llccc}
\toprule
Backbone & Dataset (Metric) & Mean $\pm$ Std & Per-seed values \\
\midrule
\multirow{5}{*}{Gemma2-2B}
 & AGNews (Acc)     & $87.76 \pm 1.62$ & 89.06, 88.28, 85.94 \\
 & REST14 (Acc)     & $76.43 \pm 2.60$ & 77.73, 73.44, 78.12 \\
 & LAP14  (Acc)     & $79.41 \pm 2.43$ & 82.07, 77.32, 78.83 \\
 & EMOC   (Acc)     & $69.46 \pm 1.07$ & 68.36, 70.51, 69.53 \\
 & CommonGen (BLEU-4) & $10.47 \pm 0.62$ & 11.09, 10.47, 9.85 \\
\midrule
\multirow{6}{*}{Llama3.1-8B}
 & AGNews (Acc)     & $87.44 \pm 0.69$ & 88.09, 86.72, 87.50 \\
 & REST14 (Acc)     & $79.89 \pm 1.69$ & 81.84, 78.91, 78.91 \\
 & LAP14  (Acc)     & $81.28 \pm 0.76$ & 82.07, 80.56, 81.21 \\
 & EMOC   (Acc)     & $72.13 \pm 1.81$ & 71.09, 74.22, 71.09 \\
 & CommonGen (BLEU-4) & $11.07 \pm 0.53$ & 10.90, 10.65, 11.67 \\
 & GSM8K  (EM)      & $55.20 \pm 0.91$ & 54.66, 56.25, 54.69 \\
\bottomrule
\end{tabular}
\end{table}

The results show two consistent patterns. First, PULSE-Retriever is
substantially more stable on Llama3.1-8B than on Gemma2-2B: standard
deviations on Llama3.1-8B range from $0.53$ to $1.81$ across all six
benchmarks, whereas on Gemma2-2B they range from $0.62$ to $2.60$. We
attribute this gap primarily to the smaller capacity of Gemma2-2B, which
makes the identified utility-localization vector more sensitive to the
particular discovery-set composition sampled by each seed. Second, on
Llama3.1-8B the per-seed values for every benchmark fall within roughly one
standard deviation of the mean, and the GSM8K result in particular is highly
stable (std $0.91$ over a $+3.16$-point reported gain), supporting the
robustness of the reasoning-task improvements reported in
Section~\ref{sec:results}.

We note two limitations of this analysis. First, three seeds yield only a
coarse variance estimate, and a larger seed budget would tighten the
confidence intervals reported above. Second, the seed sweep here varies only
the PULSE-Retriever discovery and candidate sampling; baselines in the main
tables are reported under their own default protocols. A fully matched
multi-seed comparison across all methods, together with paired significance
tests on per-query outcomes, is an important direction for further
strengthening the empirical claims and is left to future work.

\section{Limitations}
\label{Limitations}
PULSE relies on reliable SAEs for the target backbone and a small labeled
discovery split, although it is gradient-free and does not train a retriever.
It also adds offline compute and storage cost, since feature discovery requires
full-prompt backbone and SAE forward passes over multiple candidate sets, and
PULSE-Retriever caches zero-shot SAE encodings for the support pool. These costs
avoid exhaustive scoring over all $k$-shot subsets, but remain higher than
lightweight embedding retrievers.

PULSE-Retriever is a scalable proxy rather than an exact set-utility
decomposition: the utility-localization vector is estimated from set-level
activation shifts but used as a per-demonstration magnitude mask for retrieval.
Moreover, mean-pooled SAE activations may be affected by prompt-length
differences, suggesting future extensions such as query-segment pooling,
length-matched sets, or length-residualized activation differences.

For GSM8K-style reasoning, our discovery utility uses teacher-forced likelihood
of supervised solution sequences, which assumes access to reasoning traces.
Settings with only final answers may require alternative proxies such as
final-answer likelihood, execution-based rewards, or generated rationales.
Finally, our feature analyses are post-hoc and associational; causal
interventions on identified features remain future work.

\clearpage

\end{document}